\documentclass[10pt,journal,compsoc]{IEEEtran}
\ifCLASSOPTIONcompsoc
  \usepackage[nocompress]{cite}
\else
  \usepackage{cite}
\fi
\ifCLASSINFOpdf
\else
\fi
\usepackage{url}
\usepackage{amsmath}
\usepackage{amssymb}
\usepackage{multirow}
\usepackage{booktabs}
\usepackage{listings}
\usepackage{algorithm} 
\usepackage{graphicx}
\usepackage{color}
\usepackage{subcaption}
\usepackage{makecell}

\usepackage{enumitem}
\usepackage{ragged2e} % for justifying
\usepackage[pagebackref=false,breaklinks=true,colorlinks,bookmarks=false]{hyperref}

\begin{document}
\title{CrossFormer++: A Versatile Vision Transformer Hinging on Cross-scale Attention}

\author{Wenxiao Wang$^\dagger$,
        Wei Chen$^\dagger$,
        Qibo Qiu,
        Long Chen,
        Boxi Wu,
        Binbin Lin$^*$, \\
        Xiaofei He,~\IEEEmembership{Senior Member,~IEEE,}
        and Wei Liu$^*$,~\IEEEmembership{Fellow,~IEEE}
\IEEEcompsocitemizethanks{
% \IEEEcompsocthanksitem M. Shell was with the Department
% of Electrical and Computer Engineering, Georgia Institute of Technology, Atlanta,
% GA, 30332.\protect\\
% E-mail: see http://www.michaelshell.org/contact.html
% \IEEEcompsocthanksitem J. Doe and J. Doe are with Anonymous University.}% <-this % stops an unwanted space
\IEEEcompsocthanksitem $^\dagger$ means equal contributions, and $^*$ indicates the corresponding authors.
\IEEEcompsocthanksitem Wenxiao Wang, Boxi Wu, and Binbin Lin (email: binbinlin@zju.edu.cn) are with Zhejiang University.
\IEEEcompsocthanksitem Wei Chen and Xiaofei He are with State Key Lab of CAD$\&$CG, Zhejiang University.
\IEEEcompsocthanksitem Qibo Qiu is with Zhejiang Lab and Zhejiang University.
\IEEEcompsocthanksitem Long Chen is with Hong Kong University of Science and Technology.
\IEEEcompsocthanksitem Wei Liu (email: wl2223@columbia.edu) is with Data Platform, Tencent.}%
\thanks{Manuscript received November 29, 2023.}}

\newcommand\ie{\textit{i.e.}}
\newcommand\eg{\textit{e.g.}}

\IEEEtitleabstractindextext{

\justifying 

\begin{abstract}
While features of different scales are perceptually important to visual inputs, existing vision transformers do not yet take advantage of them explicitly.
To this end, we first propose a cross-scale vision transformer, CrossFormer. It introduces a cross-scale embedding layer (CEL) and a long-short distance attention (LSDA). On the one hand, CEL blends each token with multiple patches of different scales, providing the self-attention module itself with cross-scale features. On the other hand, LSDA splits the self-attention module into a short-distance one and a long-distance counterpart, which not only reduces the computational burden but also keeps both small-scale and large-scale features in the tokens. Moreover, through experiments on CrossFormer, we observe another two issues that affect vision transformers' performance, \ie, the enlarging self-attention maps and amplitude explosion. Thus, we further propose a progressive group size (PGS) paradigm and an amplitude cooling layer (ACL) to alleviate the two issues, respectively. The CrossFormer incorporating with PGS and ACL is called CrossFormer++. Extensive experiments show that CrossFormer++ outperforms the other vision transformers on image classification, object detection, instance segmentation, and semantic segmentation tasks. The code will be available at: https://github.com/cheerss/CrossFormer.
\end{abstract}

\begin{IEEEkeywords}
Vision Transformer, Image Classification, Object Detection, Semantic Segmentation
\end{IEEEkeywords}}

\maketitle

\IEEEdisplaynontitleabstractindextext

% \fi
\IEEEpeerreviewmaketitle

\IEEEraisesectionheading{\section{Introduction}\label{sec:introduction}}


\IEEEPARstart{T}{ransformer} based vision backbones have achieved great success in many computer vision tasks such as image classification, object detection, semantic segmentation, etc. Compared with convolutional neural networks (CNNs), vision transformers enable long range dependencies by introducing a self-attention module, endowing the models with higher capacities.

A transformer requires a sequence of tokens (\eg, word embeddings) as input. To adapt this requirement to typical vision tasks, most existing vision transformers~\cite{DBLP:conf/iclr/DosovitskiyB0WZ21,DBLP:conf/icml/TouvronCDMSJ21,DBLP:journals/corr/abs-2102-12122,DBLP:journals/corr/abs-2103-14030} produce tokens by splitting an input image into equally-sized patches.
For example, a $224 \times 224$ image can be split into $56 \times 56$ patches of size $4 \times 4$, and these patches are embedded through a linear layer to yield a token sequence.
Inside a certain transformer, self-attention is engaged to build the interactions between any two tokens.
Thus, the computational or memory cost of the self-attention module is $O(N^2)$, where $N$ is the length of a token sequence. Such a cost is too big for a visual input because its token sequence is much longer than that of NLP tasks. Therefore, the recently proposed vision transformers~\cite{DBLP:journals/corr/abs-2102-12122,DBLP:journals/corr/abs-2103-14030,DBLP:journals/corr/abs-2106-05786} develop multiple substitutes (\eg, Swin~\cite{DBLP:journals/corr/abs-2103-14030} restricts the self-attention in a small local region instead of a global region) to approximate the vanilla self-attention module with a lower cost.

Though the aforementioned vision transformers have made progress, they still face challenges in utilizing visual features of different scales, whereas multi-scale features are very vital for a lot of vision tasks. Particularly, an image often contains many objects of different sizes, and to fully understand the image, the model is required to extract features at different scales (\ie, different ranges and sizes). Existing vision transformers fail to deal with the above case due to two reasons:
(1) Those models' input tokens are generated from equally-sized patches. Though these patches theoretically have a chance to extract any scale features if only the patch size is large enough, it is difficult to promise that they can learn appropriate multi-scale features automatically in practice.
(2) Some vision transformers such as Swin~\cite{DBLP:journals/corr/abs-2103-14030} restrict the attention in a small local region, giving up the long range attention.

In this paper, we propose a cross-scale embedding layer (CEL) and a long-short distance attention (LSDA) to fill the cross-scale gap from two perspectives:

\textbf{Cross-scale Embedding Layer (CEL).} Following PVT~\cite{DBLP:journals/corr/abs-2102-12122}, we also employ a pyramid structure for our designed transformer, which naturally splits the vision transformer model into multiple stages. CEL appears at the start of each stage, which receives last stage’s output (or an input image) as input and samples patches with multiple kernels of different scales (\eg, 4×4 or 8 × 8). Then, each token is constructed by embedding and concatenating these patches. Through this way, we enforce some dimensions (\eg, from 4 × 4 patches) to focus on small-scale features only, while others (\eg, from 8 × 8 patches) have a chance to learn large-scale features, leading to a token with explicit cross-scale features.

\textbf{Long-Short Distance Attention (LSDA).} Instead of using a shifted window based self-attention like Swin, we argue that both short distance and long distance attentions are imperative for a visual input and thereby propose our LSDA. In particular, we split the self-attention module into a Short Distance Attention (SDA) and a Long Distance Attention (LDA). SDA builds the dependencies among neighboring embeddings, while LDA takes charge of the dependencies among embeddings far away from each other. SDA and LDA appear alternately in consecutive layers of a CrossFormer, which also reduce the cost of the self-attention module while keep both the short distance and long distance attentions.

% \begin{figure}[tb]
%     \centering
%     \includegraphics[width=1.0\linewidth]{fig_intro.pdf}
%     \caption{}
%     \label{fig:intro}
% \end{figure}

\textbf{Dynamic Position Bias (DPB).} Besides, following prior work~\cite{DBLP:conf/naacl/ShawUV18,DBLP:journals/corr/abs-2103-14030}, we employ a relative position bias for tokens’ position representations. The Relative Position Bias (RPB) only supports ﬁxed image/group size. However, image size for many vision tasks such as object detection is variable, so does group size for many architectures, including ours. To make the RPB more ﬂexible, we further introduce a trainable module called Dynamic Position Bias (DPB), which receives two tokens’ relative distance as input and outputs their position bias. The DPB module is optimized end-to-end in the training phase, inducing an ignorable cost but making RPB apply to variable group size.

Armed with the above three modules, we name our architecture as CrossFormer and design four variants of different depths and channels. While these variants achieve great performance, we also find that a further expansion of CrossFormer does not bring a persistent accuracy gain, but even a degradation. To this end, we analyze the output of each layer (\ie, the self-attention maps and the MLPs) and observe another two issues which affect the models' performance, \ie, the enlarging self-attention maps and amplitude explosion. Thus, we further propose a progressive group size (PGS) and an amplitude cooling layer (ACL) to alleviate the problems.

% Armed with the above three modules, we name our architecture as CrossFormer and design four variants of different depth and channels. They achieve state-of-the-art performance on representative downstream tasks. After CrossFormer, many vision transformers with different attention mechanisms are 

\textbf{Progressive Group Size (PGS).} In terms of the self-attention maps, we observe that tokens at shallow layers always attend to tokens around themselves. Whereas tokens at deep layers pay nearly equal attention to all other tokens. This phenomenon shows that vision transformers perform similarly as CNN, \ie, extracting local features at shallow layers and global features at deep layers, respectively. It will hinder the models' performance if adopting a fixed group size for all stages like most existing vision transformers. Hence, we propose to enlarge the group size progressively (PGS) from shallow to deep layers and implement a manually designed group size policy under this guide. Though it is a simple policy, PGS is actually a general paradigm. We hope to present its importance and appeal other researchers to explore more automated and adaptive group size policies.

\textbf{Amplitude Cooling Layer (ACL).} In the vision transformers, the activation's amplitude grows dramatically as the layer goes deeper. For example, in a CrossFormer-B, the maximal amplitude in the 22$_{nd}$ layer is 300 times larger than that in the 1$_{st}$ layer.
The tremendous amplitude discrepancy among layers makes the training process unstable. However, we also find that our proposed CEL can effectively suppress the amplitude. Considering that it is a bit cumbersome to put more CELs into a CrossFormer, we design a CEL-like but more lightweight layer dubbed Amplitude Cooling Layer (ACL). ACL is inserted after some CrossFormer blocks to cool down the amplitude.

We improve CrossFormer by introducing PGS and ACL, yielding CrossFormer++, and propose four new variants. Extensive experiments on four downstream tasks (\ie, image classification, object detection, semantic segmentation, and instance segmentation) show that CrossFormer++ outperforms CrossFormer and other existing vision transformers on all these tasks.

The following sections are organized as follows: the background and related works will be introduced in Sec.~\ref{sec:2}. Then, we will retrospect CrossFormer\footnote{An earlier version of this paper has appeared in ICLR 2022: \url{https://openreview.net/pdf?id=_PHymLIxuI}}~\cite{wang2021crossformer} and its CEL, LSDA, and DPB in Sec.~\ref{sec:3}. PGS, ACL, and CrossFormer++ are introduced in detail in Sec.~\ref{sec:4}. Thereafter, the experiments will be shown in Sec.~\ref{sec:5}. Finally, we will present our conclusions and future work in Sec.~\ref{sec:6}.

\section{Background and Related Works} \label{sec:2}

\textbf{Vision Transformers.} Motivated by the great success achieved by transformers in NLP, researchers have tried to design specific visual transformers for vision tasks to take advantage of the powerful attention mechanism. Early vision transformers like ViT~\cite{DBLP:conf/iclr/DosovitskiyB0WZ21} and DeiT~\cite{DBLP:conf/icml/TouvronCDMSJ21} transfer the original transformer to vision tasks, achieving impressive results and demonstrating great potentials. T2T-ViT~\cite{DBLP:journals/corr/abs-2101-11986} and VOLO~\cite{DBLP:journals/pami/YuanHJFY23} inherit the ViT's architecture while improving its input tokens. They introduce locality into tokens, which is more suitable for visual input. Furthermore, VOLO also incorporates Token Labeling~\cite{DBLP:conf/nips/JiangHYZSJWF21} and achieves state-of-the-art results on several downstream vision tasks. Later works like PVT~\cite{DBLP:journals/corr/abs-2102-12122}, Swin~\cite{DBLP:journals/corr/abs-2103-14030}, and MViTv2~\cite{Li2021MViTv2IM} combine the pyramid structure with the transformer and remove the class token used in the original architecture, enabling the use in further vision tasks like object detection and image segmentation. As is mentioned in MViTv2~\cite{Li2021MViTv2IM}, such a pyramid structure is partly inspired by CNN, which hierarchically expands the channel capacity while reducing the spatial resolution. Our work proposes that other insights from CNN may also be valuable for transformers. To be specific, the lower layers in the network tend to refine local features while the higher layers focus more on global information communications.

\textbf{Self-supervised ViTs.} In addition to architectural design, another field of ViT is exploring the self-supervised pre-training scheme to enhance its performance. Most self-supervised ViT fall into two categories: masked image modeling and contrastive learning. Masked image modeling such as MAE~\cite{DBLP:conf/cvpr/HeCXLDG22}, data2vec~\cite{DBLP:conf/icml/BaevskiHXBGA22}, and CAE~\cite{DBLP:journals/corr/abs-2202-03026} gives the model an masked image, and the ViT learns to predict the masked parts (pixels or hidden representations). Some papers also adapts masked image modeling to pyramid ViTs. The belief behind masked image modeling is that the model can predict the masked parts only if it understands the image and extracts good features. While contrastive learning~\cite{DBLP:conf/iccv/CaronTMJMBJ21,DBLP:journals/corr/abs-2304-07193} generates different views for each image. The model is training to predict whether these views are from the same or different image. Self-supervised pretraining is orthogonal to architectural design and they can be combined to further improve the ViT's performance.

\begin{figure*}[tb]
    \centering
    \includegraphics[width=1.0\linewidth]{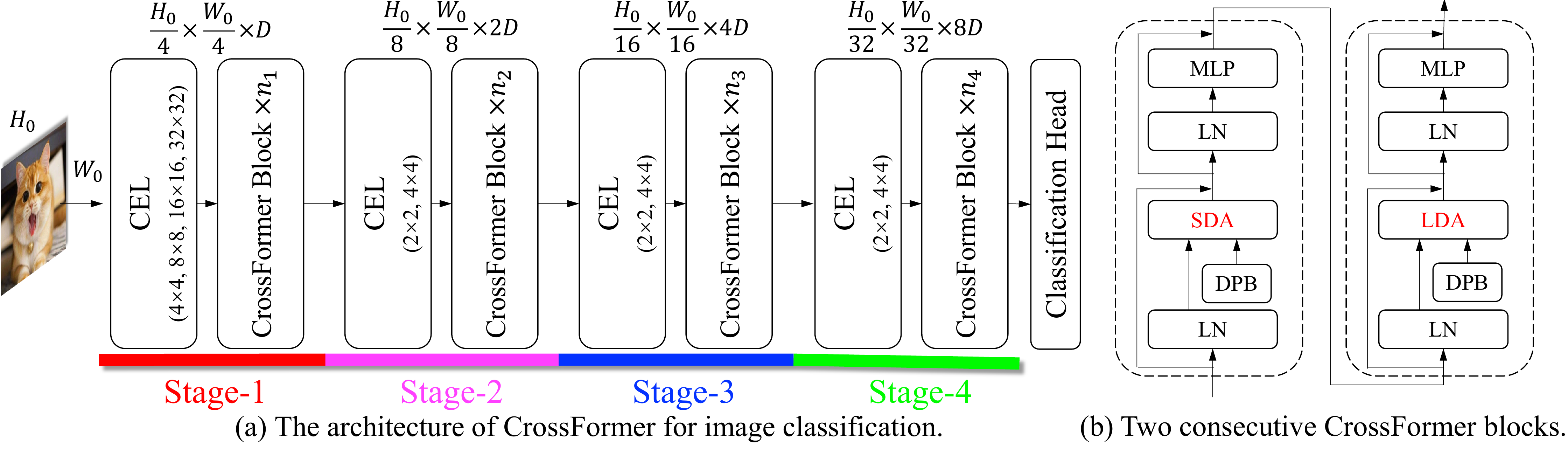}
    \caption{(a) The architecture of CrossFormer for classification. The input size is $H_0 \times W_0$, and the size of feature maps in each stage is shown on the top. \textit{Stage-i} consists of a CEL and $n_i$ CrossFormer blocks. Numbers in CELs represent kernels' sizes used for sampling patches. (b) The inner structure of two consecutive CrossFormer blocks. SDA and LDA appear alternately in different blocks.}
    \label{fig:arch}
\end{figure*}

\textbf{Efficient Self-Attention and Universal Vision Backbone.} Vanilla self-attention module in the transformer suffers from a quadratic complexity with respect to the image size, which is unacceptable for dense vision tasks like object detection and image segmentation. Therefore, in order to build transformer-based universal vision backbones, researchers have proposed ways to do efficient self-attention. The first choice is to do sparse self-attention. Instead of doing self-attention among all embeddings, some transformers divide the embeddings into groups and do self-attention within each group. The ways of choosing the group while also enabling global information interactions become the core design for works like Swin~\cite{DBLP:journals/corr/abs-2103-14030}, CAT~\cite{DBLP:journals/corr/abs-2106-05786}, CvT~\cite{DBLP:journals/corr/abs-2103-15808}, and CSwin~\cite{dong2022cswin}. Another choice is to reduce the cost of global self-attention through reducing the size of input queries and keys. PVT~\cite{DBLP:journals/corr/abs-2102-12122} uses average pooling to reduce keys size, while Scalable ViT~\cite{DBLP:journals/corr/abs-2103-10619} uses convolution with large intervals. MViTv2~\cite{Li2021MViTv2IM} inserts a pooling layer after the linear transformation layer of the self-attention module. It should also be noted that the results reported in the above works are sometimes obtained under different settings such as different depths and embedding methods, which disables the direct comparison between different ways to divide the group. Our work proposes a novel way of group division and uses the same experiment setting and architecture to compare the effectiveness of different division ways.

\textbf{Position Representations.} Transformers are permutation invariant models, which are unsatisfying for both NLP and vision tasks, where the permutation of the embeddings affects the semantic information. To make the model aware of position information, many different position representations are proposed. For example, Dosovitskiy et al. (2021)~\cite{DBLP:conf/iclr/DosovitskiyB0WZ21} directly added the embeddings with the vectors that contain absolute position information, while Relative Position Bias (RPB) (Shaw et al., 2018)~\cite{DBLP:conf/naacl/ShawUV18} shows that relative position information is more important for vision tasks and resorts to position information indicating the relative distance of two embeddings. In contrast, MaxViT~\cite{DBLP:journals/corr/abs-2204-01697} proposes that depth-wise convolutions can also be seen as a kind of conditional position representations, thus no explicit position representation is needed. Besides, Xiangxiang Chu et al.~\cite{chu2021conditional} pointed out that a successful positional encoding for vision tasks should meet the requirements that: (1) being permutation-variant but translation-invariant; (2) being able to handle different lengths of inputs; (3) containing absolute position information. Based on such requirements, they further proposed conditional positional encoding (CPE) that is dynamically generated and conditioned on the local neighborhood of the input tokens. CPE is used in latest transformers like MaxViT. Furthermore, since it could be implemented by a depthwise convolution layer, CPE is able to be combined with other designs like Squeeze-and-Excitation layer to remove explicit positional encodings layers.

\textbf{Neural Network Design.} There have been a number of works in CNN aimed at designing architectures that achieve a good trade-off between efficiency and accuracy. Works such as X3D~\cite{feichtenhofer2020x3d} have proposed greedy methods to search for good hyperparameters like spatial resolution and depth for each stage of a CNN, which attach importance to the design choice other than a novel convolution design. However, to the best of our knowledge, such an exploration in vision transformers is clearly insufficient. AutoFormer~\cite{AutoFormer} and S3 \cite{chen2021searching} (also known as AutoFormerV2) uses Neural Architecture Search (NAS) methods to find good embedding dimension for each self-attention layer. GLiT~\cite{chen2021glit} searches a good permutation of global and local self-attention modules in the transformer. Anyway, most works that focus on transformer architecture design fix their group size as 7 in order to keep pace with Swin~\cite{DBLP:journals/corr/abs-2103-14030}. Such choices are made out of convenience and clearly not the best. Our work makes a small step forward through pointing out that choosing good group size for embeddings allows the transformers to achieve a better efficiency-accuracy tradeoff.

\begin{figure*}[t]
    \centering
    \includegraphics[width=0.9\linewidth]{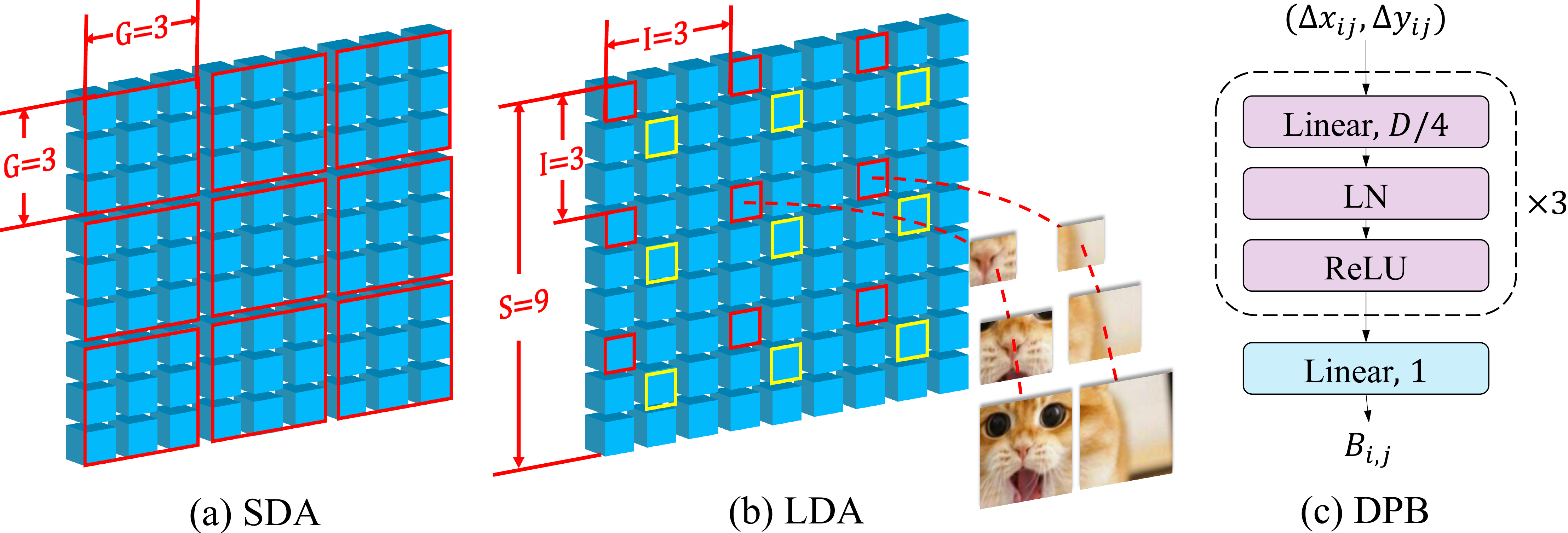}
    \caption{
        (a) Short distance attention (SDA). Embeddings (blue cubes) are grouped by red boxes.
        (b) Long distance attention (LDA). Embeddings with the same color borders belong to the same group. Large patches of embeddings in the same group are adjacent.
        (c) Dynamic position bias (DBP). The dimensions of intermediate layers are $D/4$, and the output is a scalar.}
    \label{fig:LSDA}
\end{figure*}

\begin{figure}[]
    \includegraphics[width=1.0\linewidth]{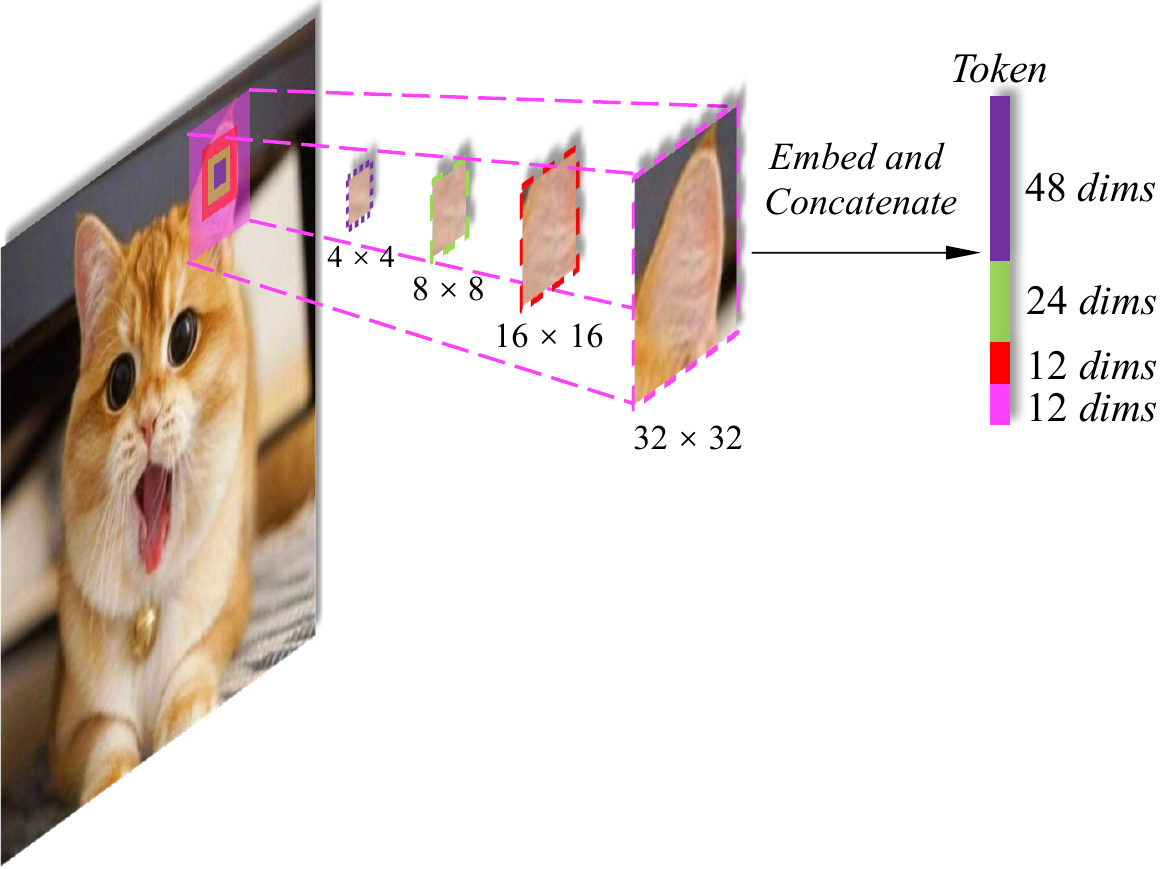}
    \hspace{-50mm}
    \renewcommand\arraystretch{1.25}
    \scalebox{0.8}{\setlength{\tabcolsep}{2.8mm}{\small
            \begin{tabular}[b]{cccc}
                \multicolumn{4}{c}{Embedding Layer} \\
                \toprule
                Type & Kernel & Stride & Dim \\
                \midrule
                Conv. & $4\times4$ & $4\times4$ & $\frac{D_t}{2}$\\
                Conv. & $8\times8$ & $4\times4$ & $\frac{D_t}{4}$ \\
                Conv. & $16\times16$ & $4\times4$ & $\frac{D_t}{8}$ \\
                Conv. & $32\times32$ & $4\times4$ & $\frac{D_t}{8}$ \\
                \bottomrule
    \end{tabular}}}
    \caption{Illustration of the CEL layer. The input image is sampled by four different kernels (\ie, $4 \times 4, 8 \times 8, 16 \times 16, 32 \times 32$) with same stride $4\times4$. Each embedding is constructed by embedding and concatenating the four patches. $D_t$ means the total dimension of the embedding.}
    \label{fig:embedding}
\end{figure}

\section{CrossFormer} \label{sec:3}

\begin{table*}[]
    \centering
    \caption{Variants of CrossFormer and CrossFormer++ for image classification. The classification head just contains a global average pooling and a linear layer, which are not shown in the table below for simplicity. $D$, $G$, and $I$ are token dimension, group size, and interval for SDA and LDA, respectively.}
    \scalebox{0.8}{
        \setlength{\tabcolsep}{3mm}{
            \begin{tabular}{c|c|c|ccccc}
                \toprule
                Version & Stage & Layer Name & Tiny (-T) & Small (-S) & Base (-B) & Large (-L) & Huge (-H) \\ 
                \midrule
                
                \multirow{20}{*}{CrossFormer} & \multirow{5}{*}{Stage-1} & Cross Embed. & \multicolumn{5}{c}{Kernel size: $4 \times 4$, $8 \times 8$, $16 \times 16$, $32 \times 32$, Stride=$4$} \\
                \cmidrule{3-8}
                & & \makecell{SDA/LDA \\ MLP} & $\begin{bmatrix}D_1=64 \\G_1=7 \\I_1=8\end{bmatrix} \times 1$ & $\begin{bmatrix}D_1=96 \\G_1=7 \\I_1=8\end{bmatrix} \times 2$ & $\begin{bmatrix}D_1=96\\G_1=7\\I_1=8\end{bmatrix} \times 2$ & $\begin{bmatrix}D_1=128 \\ G_1=7 \\ I_1=8\end{bmatrix} \times 2$ & $-$\\
                \cmidrule{2-8}
                & \multirow{5}{*}{Stage-2} & Cross Embed. & \multicolumn{5}{c}{Kernel size: $2 \times 2$, $4 \times 4$, Stride=$2$} \\
                \cmidrule{3-8}
                & & \makecell{SDA/LDA \\ MLP} & $\begin{bmatrix}D_2=128\\ G_2=7\\I_2= 4 \end{bmatrix} \times 1$ & $\begin{bmatrix}D_2=192\\G_2=7\\I_2= 4 \end{bmatrix} \times 2$ & $\begin{bmatrix}D_2=192\\G_2=7\\I_2=4 \end{bmatrix} \times 2$ & $\begin{bmatrix}D_2=256\\G_2=7\\I_2=4 \end{bmatrix} \times 2$ & $-$ \\
                \cmidrule{2-8}
                & \multirow{5}{*}{Stage-3} & Cross Embed. & \multicolumn{5}{c}{Kernel size: $2 \times 2$, $4 \times 4$, Stride=$2$} \\
                \cmidrule{3-8}
                & & \makecell{SDA/LDA \\ MLP} & $\begin{bmatrix}D_3=256 \\ G_3=7\\I_3=2 \end{bmatrix} \times 8$ & $\begin{bmatrix}D_3=384\\ G_3=7\\I_3=2 \end{bmatrix} \times 6$ & $\begin{bmatrix}D_3=384\\G_3=7\\I_3=2 \end{bmatrix} \times 18$ & $\begin{bmatrix}D_3=512\\G_3=7\\I_3=2 \end{bmatrix} \times 18$ & $-$ \\
                \cmidrule{2-8}
                & \multirow{5}{*}{Stage-4} & Cross Embed. & \multicolumn{5}{c}{Kernel size: $2 \times 2$, $4 \times 4$, Stride=$2$} \\
                \cmidrule{3-8}
                & & \makecell{SDA/LDA \\ MLP} & $\begin{bmatrix}D_4=512\\ G_4=7\\I_4=1 \end{bmatrix} \times 6$ & $\begin{bmatrix}D_4=768\\G_4=7\\I_4=1 \end{bmatrix} \times 2$ & $\begin{bmatrix}D_4=768\\G_4=7\\I_4=1 \end{bmatrix} \times 2$ & $\begin{bmatrix}D_4=1024\\ G_4=7\\I_4=1 \end{bmatrix} \times 2$ & $-$ \\
                
                \midrule

                \multirow{20}{*}{CrossFormer++} & \multirow{5}{*}{Stage-1} & Cross Embed. & \multicolumn{5}{c}{Kernel size: $4 \times 4$, $8 \times 8$, $16 \times 16$, $32 \times 32$, Stride=$4$} \\
                \cmidrule{3-8}
                & & \makecell{SDA/LDA \\ MLP/ACL} & $-$ & $\begin{bmatrix}D_1=64 \\G_1=4 \\I_1=4\end{bmatrix} \times 2$ & $\begin{bmatrix}D_1=96 \\G_1=4 \\I_1=4\end{bmatrix} \times 2$ & $\begin{bmatrix}D_1=128\\G_1=4\\I_1=4\end{bmatrix} \times 2$ & $\begin{bmatrix}D_1=128 \\ G_1=4 \\ I_1=4\end{bmatrix} \times 6$ \\
                \cmidrule{2-8}
                & \multirow{5}{*}{Stage-2} & Cross Embed. & \multicolumn{5}{c}{Kernel size: $2 \times 2$, $4 \times 4$, Stride=$2$} \\
                \cmidrule{3-8}
                & & \makecell{SDA/LDA \\ MLP/ACL} & $-$ & $\begin{bmatrix}D_2=128\\ G_2=4\\I_2= 4 \end{bmatrix} \times 2$ & $\begin{bmatrix}D_2=192\\G_2=4\\I_2= 4 \end{bmatrix} \times 2$ & $\begin{bmatrix}D_2=256\\G_2=4\\I_2=4 \end{bmatrix} \times 2$ & $\begin{bmatrix}D_2=256\\G_2=4\\I_2=4 \end{bmatrix} \times 6$ \\
                \cmidrule{2-8}
                & \multirow{5}{*}{Stage-3} & Cross Embed. & \multicolumn{5}{c}{Kernel size: $2 \times 2$, $4 \times 4$, Stride=$2$} \\
                \cmidrule{3-8}
                & & \makecell{SDA/LDA \\ MLP/ACL} & $-$ & $\begin{bmatrix}D_3=256 \\ G_3=14\\I_3=1 \end{bmatrix} \times 18$ & $\begin{bmatrix}D_3=384\\ G_3=14\\I_3=1 \end{bmatrix} \times 18$ & $\begin{bmatrix}D_3=512\\G_3=14\\I_3=1 \end{bmatrix} \times 18$ & $\begin{bmatrix}D_3=512\\G_3=14\\I_3=1 \end{bmatrix} \times 18$ \\
                \cmidrule{2-8}
                & \multirow{5}{*}{Stage-4} & Cross Embed. & \multicolumn{5}{c}{Kernel size: $2 \times 2$, $4 \times 4$, Stride=$2$} \\
                \cmidrule{3-8}
                & & \makecell{SDA/LDA \\ MLP/ACL} & $-$ & $\begin{bmatrix}D_4=512\\ G_4=7\\I_4=1 \end{bmatrix} \times 2$ & $\begin{bmatrix}D_4=768\\G_4=7\\I_4=1 \end{bmatrix} \times 2$ & $\begin{bmatrix}D_4=1024\\G_4=7\\I_4=1 \end{bmatrix} \times 2$ & $\begin{bmatrix}D_4=1024\\ G_4=7\\I_4=1 \end{bmatrix} \times 2$ \\
                
                \bottomrule
    \end{tabular}}}
    \label{tab:variants}
\end{table*}

The overall architecture of CrossFormer is plotted in Fig.~\ref{fig:arch}.
Following \cite{DBLP:journals/corr/abs-2102-12122,DBLP:journals/corr/abs-2103-14030,DBLP:journals/corr/abs-2106-05786}, CrossFormer also employs a pyramid structure, which naturally splits the transformer model into four stages.
Each stage consists of a cross-scale embedding layer (CEL, Sec.~\ref{sec:cel}) and several CrossFormer blocks (Sec.~\ref{sec:block}).
A CEL receives last stage's output (or an input image) as input and generates cross-scale tokens through an embedding layer.
In this process, CEL (except that in \textit{Stage-1}) reduces the number of embeddings to a quarter while doubles their dimensions for a pyramid structure.
Then, several CrossFormer blocks, each of which involves long short distance attention (LSDA) and dynamic position bias (DPB), are set up after CEL.
A specialized head (\eg, the classification head in Fig.~\ref{fig:arch}) follows after the final stage accounting for a specific task.

\subsection{Cross-scale Embedding Layer (CEL)} \label{sec:cel}
Cross-scale embedding layer (CEL) is leveraged to generate input tokens for each stage.     
Fig.~\ref{fig:embedding} takes the first CEL, which is ahead of \textit{Stage-1}, as an example.
It receives an image as input, then sampling patches using four kernels of different sizes.
The stride of four kernels is kept the same so that they generate the same number of tokens\footnote{The image will be padded if necessary.}.
As we can observe in Fig.~\ref{fig:embedding}, every four corresponding patches have the same center but different scales, and all these four patches will be embedded and concatenated as one embedding.
In practice, the process of sampling and embedding can be fulfilled through four convolutional layers.

For a cross-scale token, one problem is how to set the embedded dimension of each scale.
An intuitive way is allocating the dimension equally. Take a 96-dimensional token with four kernels as an example, each kernel outputs a 24-dimensional vector and concatenating these vectors yields a 96-dimensional token. 
However, the computational budget of a convolutional layer is proportional to $K^2D^2$, where $K$ and $D$ represent kernel size and input/output dimension, respectively (assuming that the input dimension equals to the output dimension).
Therefore, given the same dimension, a large kernel consumes a greater budget than a smaller one.
To control the total budget of the CEL, we use a lower dimension for large kernels while a higher dimension for small kernels.
Fig.~\ref{fig:embedding} provides the specific allocation rule in its subtable, and a $96$-dimensional example is given.
Compared with allocating the dimension equally, our scheme saves much computational cost but does not explicitly affect the model's performance.
The cross-scale embedding layers in other stages work in a similar way. As shown in Fig.~\ref{fig:arch}, CELs for \textit{Stage-2/3/4} use two different kernels ($2 \times 2$ and $4 \times 4$). Further, to form a pyramid structure, the strides of CELs for \textit{Stage-2/3/4} are set as $2 \times 2$ to reduce the number of embeddings to a quarter.

\subsection{CrossFormer Block}  \label{sec:block}

Each CrossFormer block consists of a long short distance attention module (\ie, LSDA, which involves a short distance attention (SDA) module or a long distance attention (LDA) module) and a multi-layer perceptron (MLP).
As shown in Fig.~\ref{fig:arch}b, SDA and LDA appear alternately in different blocks, and the dynamic position bias (DPB) module works in both SDA and LDA for obtaining embeddings' position representations.
Following the prior vision transformers, residual connections are used in each block.

\subsubsection{Long Short Distance Attention (LSDA)} \label{sec:LSDA}

We split the self-attention module into two parts: short distance attention (SDA) and long distance attention (LDA).
For SDA, all $G \times G$ adjacent embeddings are grouped together. Fig.~\ref{fig:LSDA}a gives an example where $G=3$.
For LDA with input of size $S \times S$, the embeddings are sampled with a fixed interval $I$. For example in Fig.~\ref{fig:LSDA}b ($I=3$), all embeddings with a red border belong to a group, and those with a yellow border comprise another group. The group's height or width for LDA is computed as $G=S/I$ (\ie, $G=3$ in this example). After grouping embeddings, both SDA and LDA employ the vanilla self-attention within each group. As a result, the memory/computational cost of the self-attention module is reduced from $O(S^4)$ to $O(S^2G^2)$.

It is worth noting that the effectiveness of LDA also benefits from cross-scale embeddings. Specifically,
we draw all the patches comprising two embeddings in Fig.~\ref{fig:LSDA}b.
As we can see,
the small-scale patches of two embeddings are non-adjacent, so it is difficult to judge their relationship without the help of the context.
In other words, it will be hard to build the dependency between these two embeddings if they are only constructed by small-scale patches (\ie, single-scale feature). On the contrary, adjacent large-scale patches provide sufficient context to link these two embeddings, which makes long-distance cross-scale attention easier to compute and more meaningful.

% We provide the pseudo-code of LSDA in the appendix (\ref{apd:pesudo}). Based on the vanilla multi-head self-attention, LSDA can be implemented with only ten lines of code. Further, only \textit{reshape} and \textit{permute} operations are used, so no extra computational cost is introduced.

\begin{figure*}[]
    \includegraphics[width=1.0\linewidth]{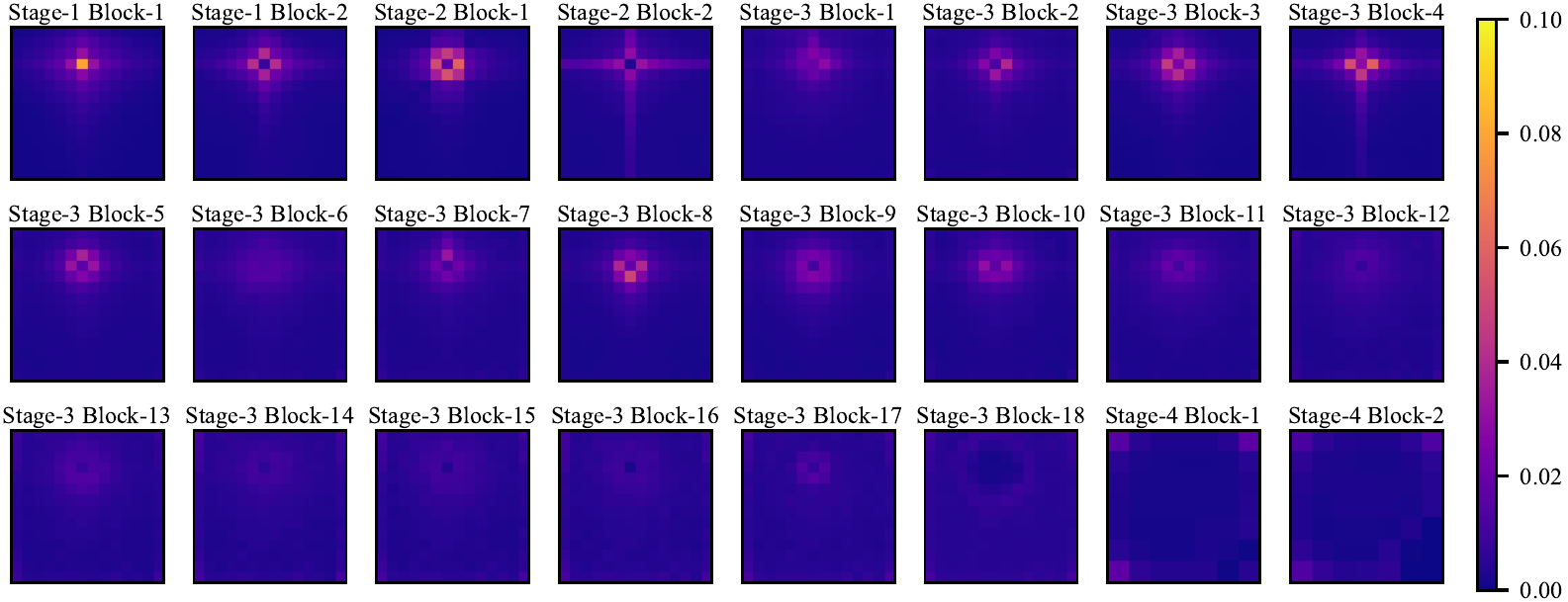}
    \caption{The attention maps of a random token in CrossFormer-B's blocks. The attention map size is $14 \times 14$ (except $7 \times 7$ for Stage-4). The attention concentrates around each token itself at shallow blocks but gradually disperses and distributes evenly at deep blocks.}
    \label{fig:mean_attn}
\end{figure*}

\begin{figure}[]
    \includegraphics[width=1.0\linewidth]{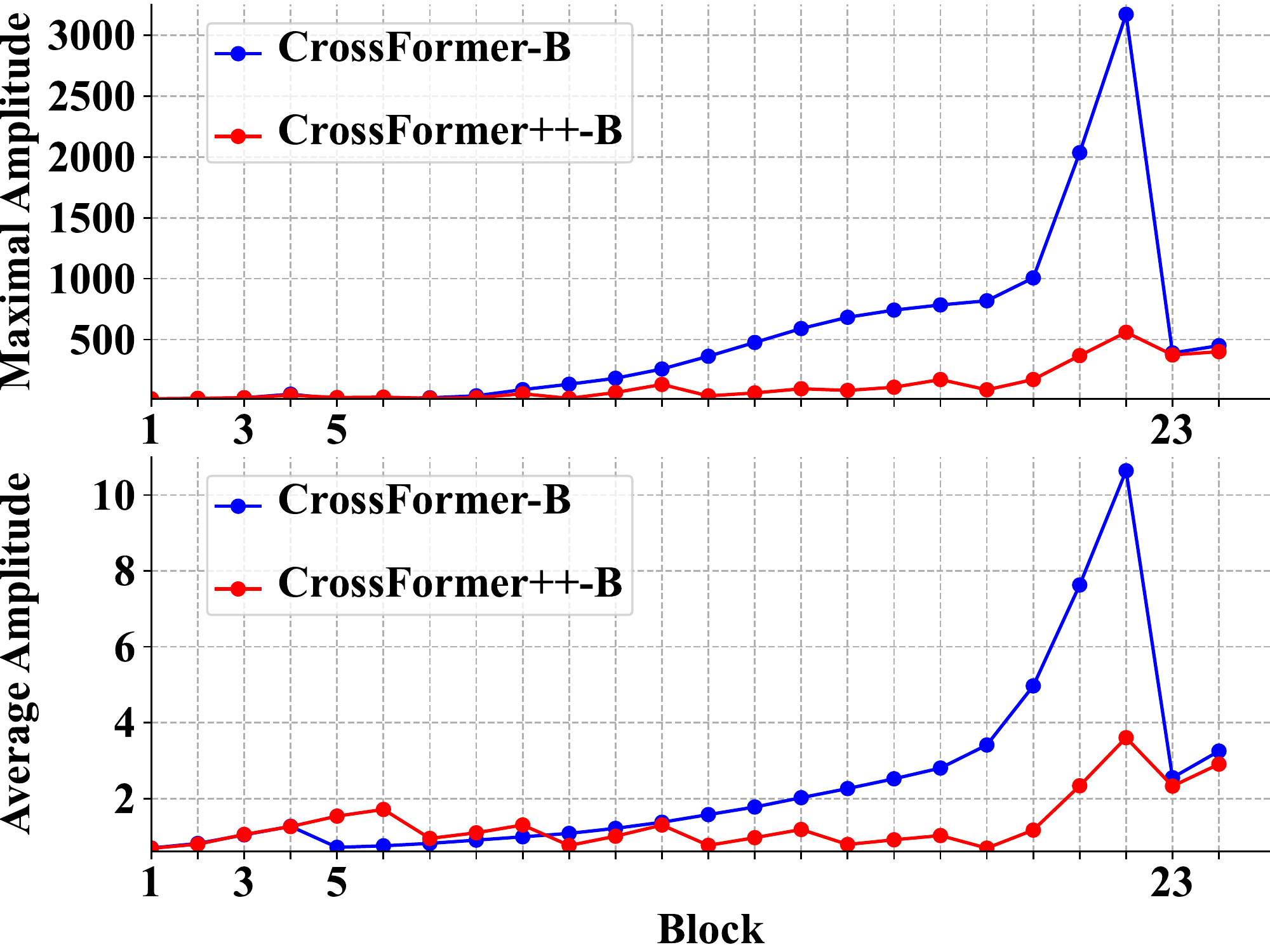}
    \caption{Maximal/Average amplitude of each block's output for CrossFormer-B and CrossFormer++-B. Both models use the same group size for all stages. CrossFormer++-B contains ACLs while CrossFormer-B does not.}
    \label{fig:amplitude}
\end{figure}

\begin{figure*}[]
    \includegraphics[width=1.0\linewidth]{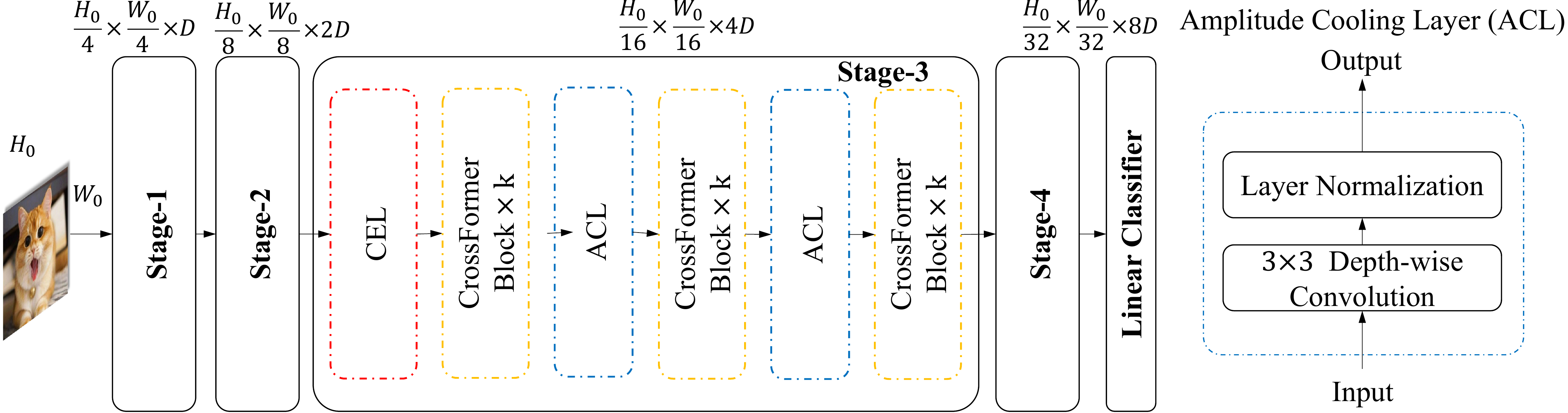}
    \caption{The architecture of CrossFormer++ and its amplitude cooling layer (ACL).}
    \label{fig:architecture++}
\end{figure*}

\subsubsection{Dynamic Position Bias (DPB)} \label{sec:dpb}

Relative position bias (RPB) indicates embeddings' relative positions by adding a bias to their attentions. Formally, the LSDA's attention map with RPB becomes:
\begin{equation}
    \label{equ:attn}
    Attn = \mathtt{Softmax}(QK^T/\sqrt d + B)V,
\end{equation}
where $Q, K, V \in \mathbb{R}^{G^2 \times D}$ represent \textit{query, key, value} in the self-attention module, respectively. $\sqrt{d}$ is a constant normalizer, and $B \in \mathbb{R}^{G^2 \times G^2}$ is the RPB matrix. In previous work~\cite{DBLP:journals/corr/abs-2103-14030}, $B_{i,j} = \hat{B}_{\Delta x_{ij}, \Delta y_{ij}}$, where $\hat{B}$ is a fixed-sized matrix, and $(\Delta x_{ij}, \Delta y_{ij})$ is the coordinate distance between the $i_{th}$ and $j_{th}$ embeddings. It is obvious that the image/group size is restricted in case that $(\Delta x_{ij}, \Delta y_{ij})$ exceeds the size of $\hat{B}$. In contrast, we propose an MLP-based module called DPB to generate the relative position bias dynamically, \ie,
\begin{equation}
    \label{equ:attn_bias}
    B_{i,j} = \mathtt{DPB}(\Delta x_{ij}, \Delta y_{ij}).
\end{equation}
The structure of DPB is displayed in Fig.~\ref{fig:LSDA}c. Its non-linear transformation consists of three fully-connected layers with layer normalization~\cite{DBLP:journals/corr/BaKH16} and ReLU~\cite{DBLP:conf/icml/NairH10}.
The input dimension of DPB is $2$, \ie, $(\Delta x_{ij}, \Delta y_{ij})$, and intermediate layers' dimension is set as $D/4$, where $D$ is the dimension of embeddings. The output $B_{ij}$ is a scalar, encoding the relative position feature between the $i_{th}$ and $j_{th}$ embeddings.
DPB is a trainable module optimized along with the whole transformer model. It can deal with any image/group size without worrying about the bound of $(\Delta x_{ij}, \Delta y_{ij})$.

In the appendix, we prove that DPB is equivalent to RPB if the image/group size is fixed. In this case, we can transform a trained DPB to RPB in the testing phase. We also provide an efficient $O(G^2)$ implementation of DPB when image/group size is variable (the complexity is $O(G^4)$ in a normal case because $B \in \mathbb{R}^{G^2 \times G^2}$).

\subsection{Variants of CrossFormer}

TABLE~\ref{tab:variants} lists the detailed configurations of CrossFormer's four variants for image classification.
To re-use the pre-trained weights, the models for other tasks (\eg, object detection) employ the same backbones as classification except that they may use different $G$ and $I$. Specifically, besides the configurations same to classification, we also test with $G_1=G_2=14, I_1 = 16$, and $I_2 = 8$ for the detection (and segmentation) models' first two stages to adapt to larger images. 
% The specific configurations are described in the appendix (\ref{apd:variants}).
Notably, the group size or the interval (\ie, $G$ or $I$) does not affect the shape of weight tensors, so the backbones pre-trained on ImageNet can be readily fine-tuned on other tasks even though they use different $G$ or $I$.

\section{CrossFormer++} \label{sec:4}
In this section, we propose a progressive group size (PGS) paradigm to adapt to vision transformers' gradually expanding group size. Moreover, an amplitude cooling layer (ACL) is proposed to alleviate the amplitude explosion issue. The PGS and ACL are plugged into CrossFormer, yielding an improved version, dubbed CrossFormer++.
% On the basis of CrossFormer, we further enlarge the CrossFormer for higher capacity models. Unfortunately, this does not bring ulterior improvement, but even accuracy degradation. We explore reasons by visualizing the attention maps and the output of MLPs. The results indicate that tokens prone to attending to local regions in shallow layers while attending to global regions in deep layers. To accomendate this, we propose Progressive Group Size (PGS), which use a small group size at early stages while enlarge it gradually at later stages. Moreover, we find the amplitude grows dramatically as the layer goes deeper. Thus, we propose an Amplitude Cooling Layer (ACL) to alleviate the amplitude explosion problem. Hinging on PGS and ACL, we further propose the improved version of CrossFormer, dubbed CrossFormer++.

\subsection{Progressive Group Size (PGS)}
Existing work~\cite{DBLP:conf/nips/RaghuUKZD21} has explored the vanilla ViTs' mechanisms and found that ViTs prefer global attentions even from early layers, which work in a different way from CNNs. However, vision transformers with a pyramid structure take smaller patches as input, resorting to group self-attention, and may perform differently from the vanilla ViTs. We take CrossFormer as an example and compute its average attention maps. Specifically, the attention maps of a certain group can be represented as:
\begin{align}
    Attn \in \mathbb{R}^{B\times H \times G \times G \times G \times G},
    % Attn_{b, h, i, j} \in \mathbb{R}^{G \times G}
\end{align}
where $B, H, G$ represent batch size, number of heads, and group size, respectively. It means that there are $G \times G$ tokens in all, and that the attention map for each token is of size $G \times G$. The attention map of a token at image $b$, head $h$, and position $(i, j)$ is represented as:
\begin{align}
    Attn_{b, h, i, j} \in \mathbb{R}^{G \times G}.
\end{align}
For the token at position $(i, j)$, the attention map averaged over batches and multi-heads is:
\begin{align}
    \overline{Attn}_{i, j} = \frac{1}{BH}\sum_{b=0}^{B}\sum_{h=0}^{H}Attn_{b, h, i,j} \in \mathbb{R}^{G \times G}.
\end{align}
We train a CrossFormer-B with a large group size of $14 \times 14$ and compute the average attention $\overline{Attn}$ of each token. The results of a random token's $\overline{Attn}$ are shown in Fig.~\ref{fig:mean_attn}. As we can see, the attention regions gradually expand from shallow to deep layers. For example, tokens in the first two stages mainly attend to regions of size $4 \times 4$ around themselves, while the attention maps of deep layers from stage-3 are evenly distributed. The results indicate that tokens from shallow layers prefer local dependencies, while those from deep layers prefer global attentions.

To this end, we propose a PGS paradigm, \ie, adopting a smaller group size in shallow layers to lower the computational budget and a larger group size in deep layers for global attentions. 
Under this guide, we first empirically set the group size to $\{4, 4, 14, 7\}$\footnote{The last stage's group size decreases because its feature maps size is $7 \times 7$, and a group size of $7$ already means a global attention.} for four stages, respectively. 
Then, a linear scaling group size policy is proposed, \ie, expanding the group size linearly from shallow to deep layers.

Previous work S3 \cite{chen2021searching} has proposed a design guideline similar to PGS. However, the intuitions behind and the details are different. The guideline from S3 is inspired by the phenomenon observed during the process of NAS. The search space of group size they use is limited to only two integers, i.e. $\{7, 14\}$, and the effect of different group sizes is modeled with linear approximation, which is relatively coarser compared with PGS. Contrarily, PGS gets intuition from attention matrix visualization and adopts a more aggressive and flexible strategy that enables different transformer blocks to choose different group size from a set of consecutive integers within interval $[4, 14]$, e.g. linear scaling group size policy. We compare experiment results and provide detailed ablation experiments to prove the effectiveness of PGS in section \ref{sec:5}. The combination of PGS and NAS is left for future research.

% \begin{equation}
%     G = \lfloor \frac{1}{\mathtt{depth}} \times S \rfloor
% \end{equation}

\subsection{Amplitude Cooling Layer (ACL)}
In addition to attention maps, we also explore the output's amplitude of each block. As shown in Fig.~\ref{fig:amplitude}, for a CrossFormer-B model, the amplitude increases greatly as the block goes deeper. In particular, the maximal output for the 22$_{nd}$ block becomes over 3000, about 300 times larger than the value for the 1$_{st}$ block. Besides, the average amplitude of the 22$_{nd}$ also becomes about 15 times larger than that of the 1$_{st}$ block. The extreme value makes the training process unstable and hinders the model from converging. 

Fortunately, we also observe that the amplitude shrinks back to a small value at the start of each stage (\eg, the $5_{th}$ and $23_{th}$ blocks). 
We think that all block's outputs are gradually accumulated through the residual connections in the model. The CEL at the beginning of each stage does not have a residual connection and cuts off the accumulation process, so it can effectively cool down the amplitude.
% Moreover, the depth of a CEL is two (a convolutional layer and a normalization layer), which is shallow enough, so it will not increase to model depth and affect the back-propagation.
While CEL still contains cumbersome normal convolution layers, we propose a more lightweight counterpart, dubbed amplitude cooling layer (ACL). As shown in Fig.~\ref{fig:architecture++}, similar to CEL, an ACL does not use any residual connection, either. It only consists of a depth-wise convolution layer and a normalization layer. The comparison in Fig.~\ref{fig:amplitude} shows that ACL can also cool down the amplitude, but it introduces fewer parameters and a less computational budget than CEL because a depth-wise convolution with a small kernel (instead of a normal convolutional layer) is used.

However, ACL without a residual connection will prolong the back-propagation path and aggravate the vanishing gradient issue. To prevent this, we put an ACL layer after each $k$ blocks with $k > 1$, as shown in Fig.~\ref{fig:architecture++}. Empirically, $k = 3$ achieves a satisfying trade-off between amplitude cooling and back-propagation.

\subsection{Variants of CrossFormer++}
Armed with PGS and ACL, we further improve CrossFormer and propose CrossFormer++. The architectures are listed as TABLE~\ref{tab:variants}.
Wherein, CrossFormer++-B and CrossFormer++-L inherit each stage's depth and channel from CrossFormer-B and CrossFormer-L, respectively.
Besides, CrossFormer++ focuses more on larger models, so the ``Tiny (-T)'' version is ignored, and a new ``Huge (-H)'' version is constructed.
Particularly, CrossFormer++-H puts more layers in the first two stages for refined low-level features. Moreover, we find that for the ``Small (-S)'' version, a deep slim model performs similar to a shallow wide one but introduces fewer parameters, so CrossFormer++-S resorts to a deeper model than CrossFormer-S while using fewer channels for each block.

\section{Experiments} \label{sec:5}

The experiments are carried out on four challenging tasks: image classification, object detection, instance segmentation, and semantic segmentation. To entail a fair comparison, we keep the same data augmentation and training settings as the other vision transformers as far as possible.
The competitors are all competitive vision transformers, including DeiT~\cite{DBLP:conf/icml/TouvronCDMSJ21}, PVT~\cite{DBLP:journals/corr/abs-2102-12122}, T2T-ViT~\cite{DBLP:journals/corr/abs-2101-11986}, TNT~\cite{DBLP:journals/corr/abs-2103-00112}, CViT~\cite{DBLP:journals/corr/abs-2103-14899}, Twins~\cite{DBLP:journals/corr/abs-2104-13840}, Swin~\cite{DBLP:journals/corr/abs-2103-14030}, S3~\cite{chen2021searching}, 
NesT~\cite{DBLP:journals/corr/abs-2105-12723}, CvT~\cite{DBLP:journals/corr/abs-2103-15808},  ViL~\cite{DBLP:journals/corr/abs-2103-15358},
CAT~\cite{DBLP:journals/corr/abs-2106-05786}, ResT~\cite{DBLP:journals/corr/abs-2105-13677}, TransCNN~\cite{DBLP:journals/corr/abs-2106-03180}, Shuffle~\cite{DBLP:journals/corr/abs-2106-03650}, BoTNet~\cite{DBLP:journals/corr/abs-2101-11605}, RegionViT~\cite{DBLP:journals/corr/abs-2106-02689}, ViTAEv2~\cite{DBLP:journals/corr/abs-2202-10108}, MPViT~\cite{DBLP:conf/cvpr/LeeKWH22}, ScalableViT~\cite{DBLP:conf/eccv/YangMWTXZL22}, DaViT~\cite{DBLP:conf/eccv/DingXCLWY22}, and CoAtNet~\cite{DBLP:conf/nips/DaiLLT21}.

\begin{table*}[t]
    \centering
    \caption{Results on the ImageNet validation set. The input size is $224 \times 224$ for most models, while it is $384 \times 384$ for the model with a $^\dagger$. Results of other architectures are drawn from original papers. The \textcolor{blue}{blue} numbers in brackets represent the improvements of CrossFormer++ over CrossFormer. Throughput is tested with batch size (bs.) being 64 or 1.}
    \scalebox{1.0}{
        \setlength{\tabcolsep}{1mm}{
            \begin{subtable}[h]{0.5\textwidth}
                \begin{tabular}{l|rrlr}
                    \toprule
                    \multirow{2}{*}{Architectures} & \multirow{2}{*}{\#Params} & \multirow{2}{*}{FLOPs} & \multirow{2}{*}{Accuracy} & Throughput \\
                    & & & & (bs. = 64 / 1 )\\
                    \midrule
                    PVT-S & 24.5M & 3.8G & 79.8\% & \ \ 915 / \ \ \textbf{82} \\
                    RegionViT-T & 13.8M & 2.4G & 80.4\% & 1017 / \ \ 50\\
                    Twins-SVT-S & 24.0M & 2.8G & 81.3\% & 1130 / \ \ \textbf{82} \\
                    \textbf{CrossFormer-T} & 27.8M & 2.9G & \textbf{81.5\%} & \textbf{1157} / \ \ 63 \\
                    \midrule
                    DeiT-S & 22.1M & 4.6G & 79.8\% & \textbf{984} / 136 \\
                    T2T-ViT & 21.5M & 5.2G & 80.7\% & 808 /  \ \ 86 \\
                    CViT-S & 26.7M & 5.6G & 81.0\% & 719 / \ \ 88 \\
                    PVT-M & 44.2M & 6.7G & 81.2\%  & 571 / \ \ 50 \\
                    TNT-S & 23.8M & 5.2G & 81.3\% & 501 / \ \ 67 \\
                    Swin-T & 29.0M & 4.5G & 81.3\% & 809 / 112 \\
                    NesT-T & 17.0M & 5.8G & 81.5\% & 627 / \textbf{145} \\
                    CvT-13& 20.0M & 4.5G & 81.6\% & 867 / \ \ 67 \\
                    ResT & 30.2M & 4.3G & 81.6\% & 830 / \ \ 95 \\
                    CAT-S & 37.0M & 5.9G & 81.8\% & 589 / \ \ 82 \\
                    ViL-S & 24.6M & 4.9G & 81.8\% &  414 / \ \ 62 \\
                    S3-T & 28.1M & 4.7G & 82.1\% &  700 / \ 106 \\
                    RegionViT-S & 30.6M & 5.3G & 82.5\% & 611 / \ \ 63 \\
                    Shuffle-T & 29.0M & 4.6G & 82.5\% & 757 / \ \ 94 \\
                    ViTAEv2-S & 19.2M & 5.7G & 82.6\% & 521 / \ \ 53 \\ 
                    CSWin-T & 23.0M & 4.3G & 82.7\% &  643 / \ \ 25 \\
                    Shunted-S & 22.4M & 4.9G & 82.9\% & 581 / \ \ 38 \\
                    MPViT-S & 22.8M & 4.7G & 83.0\% & 501 / \ \ 27 \\
                    ScalableViT-S & 32.0M & 4.2G & 83.1\% & 720 / \ \ 46 \\
                    \textbf{CrossFormer-S} & 30.7M & 4.9G & \textcolor{blue}{82.5\%} & 733 / \ \ 77\\
                    \textbf{CrossFormer++-S} & 23.3M & 4.4G & \textbf{83.2\%} (\textcolor{blue}{+0.8\%}) & 665 / \ \ 41 \\
                    \bottomrule
                \end{tabular}
        \end{subtable}}
        
        \setlength{\tabcolsep}{1mm}{
            \begin{subtable}[h]{0.57\textwidth}
                \begin{tabular}{l|rrlr}
                    \toprule
                    \multirow{2}{*}{Architectures} & \multirow{2}{*}{\#Params} & \multirow{2}{*}{FLOPs} & \multirow{2}{*}{Accuracy} & Throughput \\
                    & & & & (bs. = 64 / 1 )\\
                    \midrule
                    BoTNet-S1-59 & 33.5M & 7.3G & 81.7\% & \textbf{668} / \textbf{110} \\
                    PVT-L & 61.4M & 9.8G & 81.7\% & 397 / \ \ 33 \\ 
                    CvT-21 & 32.0M & 7.1G & 82.5\% & 559 / \ \ 42 \\ 
                    CAT-B & 52.0M & 8.9G & 82.8\% & 423 / \ \ 55 \\ 
                    Swin-S & 50.0M & 8.7G & 83.0\% & 454 / \ \ 42 \\ 
                    S3-S & 50.0M & 9.5G & 83.7\% & 373 / \ \ 56 \\
                    S3-B & 73.0M & 13.6G & 84.0\% & 280 / \ \ 40 \\
                    RegionViT-M & 41.2M & 7.4G & 83.1\% & 479 / \ \ 36 \\ 
                    Twins-SVT-B  & 56.0M & 8.3G & 83.1\% & 479 / \ \ 46 \\
                    NesT-S & 38.0M & 10.4G & 83.3\% & 373 / \ \ 79 \\
                    \textbf{CrossFormer-B} & 52.0M & 9.2G & \textcolor{blue}{83.4\%} & 431 / \ \ 42 \\
                    \textbf{CrossFormer++-B} & 52.0M & 9.5G &\textbf{84.2\%} (\textcolor{blue}{+0.8\%}) & 380 / \ \ 42 \\
                    \midrule
                    DeiT-B & 86.0M & 17.5G & 81.8\% & 306 / \textbf{157} \\ 
                    DeiT-B$^\dagger$ & 86.0M & 55.4G & 83.1\% & 94 / \ \ 74\\
                    ViL-B & 55.7M & 13.4G & 83.2\% & 160 / \ \ 24 \\
                    RegionViT-B & 72.0M & 13.3G & 83.3\% & \textbf{314} / \ \ 42 \\
                    Twins-SVT-L & 99.2M & 14.8G & 83.3\% & 310 / \ \ 59 \\
                    Swin-B & 88.0M & 15.4G & 83.3\% & 295 / \ \ 55 \\
                    NesT-B & 68.0M & 17.9G & 83.8\% & 246 / \ \ 61 \\
                    DaViT-Base & 87.9M & 15.5G & 84.6\% & 279 / \ \ 56 \\
                    \textbf{CrossFormer-L} & 92.0M & 16.1G & \textcolor{blue}{84.0\%} & 285 / \ \ 41 \\
                     \textbf{CrossFormer++-L} & 92.0M & 16.6G & \textbf{84.7\%} (\textcolor{blue}{+0.7\%}) & 253 / \ \ 41 \\
                    \midrule
                    ScalableViT-L & 104.0M & 14.7G & 84.4\% & \textbf{280} / \ \ \textbf{50} \\
                    CoAtNet & 168.0M & 34.7G & 84.5\% & 183 / \ \ 44 \\
                    \textbf{CrossFormer++-H} & 96.0M & 21.8G & \textbf{84.9\%} & 179 / \ \ 31 \\
                    \bottomrule
                \end{tabular}
    \end{subtable}}}
    \label{tab:classification}
\end{table*}

\begin{table*}[]
    \centering
    \caption{Object detection results on the COCO 2017 \textit{val} set with RetinaNets as detectors. Numbers in \textcolor{blue}{blue} fonts represent the improvement of CrossFormer++ over CrossFormer. CrossFormers with $^\ddagger$ use different group sizes from classification models. FPS means frames per second.}
    \scalebox{0.97}{\setlength{\tabcolsep}{2.8mm}{
            \begin{tabular}{c|l|rr|lll|lll|r}
                \toprule
                Method & Backbone & \#Params & FLOPs & AP$^b$ & AP$^b_{50}$ & AP$^b_{75}$ & AP$^b_{S}$ & AP$^b_{M}$ & AP$^b_{L}$ & FPS \\
                \midrule
                \multirow{22}{*}{RetinaNet} & PoolFormer-S36 & 40.6M & $-$ & 39.5 & 60.5 & 41.8 & 22.5 & 42.9 & 52.4 & 6 \\ 
                \multirow{22}{*}{$1\times$ schedule} & CAT-B & 62.0M & 337.0G & 41.4 & 62.9& 43.8 & 24.9 & 44.6 & 55.2 & 12 \\
                & Swin-T & 38.5M & 245.0G  & 41.5 & 62.1 & 44.2 & 25.1 & 44.9 & 55.5 & \textbf{20} \\
                % & PVT-M & 53.9M & $-$ & 41.9 & 63.1 & 44.3 & 25.0 & 44.9 & 57.6 \\
                & ViL-M & 50.8M & 338.9G & 42.9 & 64.0 & 45.4 & 27.0 & 46.1 & 57.2 & 6 \\
                & RegionViT-B & 83.4M & 308.9G & 43.3 & 65.2 & 46.4 & 29.2 & 46.4 & 57.0 & 10\\
                & TransCNN-B & 36.5M & $-$ & 43.4 & 64.2 & 46.5 & 27.0 & 47.4 & 56.7 & 6 \\
                & DaViT-Tiny & $-$ & 244.0G & 44.0 & $-$ & $-$ & $-$ & $-$ & $-$ & 18 \\
                & PVTv2-B3 & 35.1M & $-$ & 44.6 & 65.6 & 47.6 & 27.4 & 48.8 & 58.6 & 9 \\
                & \textbf{CrossFormer-S} & 40.8M & 282.0G & 44.4 & 65.8 & 47.4 & 28.2 & 48.4 & 59.4 & 13 \\
                % & ScalableViT-S & 36.0M & 238.0G & 45.2 & 66.5 & 48.4 & 29.2 & 49.1 & 60.3 \\
                % & Shunted-S & 32.1M & $-$ & 45.4 & 65.9 & 49.2 & 28.7 & 49.3 & 60.0 \\
                & \textbf{CrossFormer-S$^\ddagger$} & 40.8M & 272.1G & 44.2 & 65.7 & 47.2 & 28.0 & 48.0 & 59.1 & 14 \\
                & \textbf{CrossFormer++-S$^\ddagger$} & 40.8M & 272.1G & \textbf{45.1}\,\textcolor{blue}{(+0.9)} & 66.6 & 48.5 & 28.7 & 49.4 & 60.3 & 16 \\
                \cmidrule{2-11}
                % & PVT-L & 71.1M & 345.0G & 42.6 & 63.7 & 45.4 & 25.8 & 46.0 & 58.4 \\
                & Twins-SVT-B & 67.0M & 322.0G & 44.4 & 66.7 & 48.1 & 28.5 & 48.9 & 60.6 & 10 \\
                & RegionViT-B+ & 84.5M & 328.2G & 44.6 & 66.4 & 47.6 & 29.6 & 47.6 & 59.0 & 8 \\
                & Swin-B & 98.4M & 477.0G & 44.7 & 65.9 & 49.2 & $-$ & $-$ & $-$ & 10 \\
                & Twins-SVT-L & 110.9M & 455.0G & 44.8 & 66.1 & 48.1 & 28.4 & 48.3 & 60.1 & 7 \\
                & ScalableViT-B & 85.0M & 330.0G & 45.8 & 67.3 & 49.2 & 29.9 & 49.5 & 61.0 & 10 \\
                & DaViT-Small & $-$ & 332.0G & 46.0 & $-$ & $-$ & $-$ & $-$ & $-$ & \textbf{12} \\
                & PVTv2-B4 & 72.3M & $-$ & 46.1 & 66.9 & 49.2 & 28.4 & 50.0 & 62.2 & 7 \\
                & \textbf{CrossFormer-B} & 62.1M & 389.0G & 46.2 & 67.8 & 49.5 & 30.1 & 49.9 & 61.8 & 9 \\
                & \textbf{CrossFormer-B$^\ddagger$} & 62.1M & 379.1G & 46.1 & 67.7 & 49.0 & 29.5 & 49.9 & 61.5 & 9 \\
                & \textbf{CrossFormer++-B$^\ddagger$} & 62.2M & 389.0G & \textbf{46.6}\,\textcolor{blue}{(+0.5)} & 68.4 & 50.1	& 31.3 & 50.8 & 61.9 & 11 \\
                \bottomrule
    \end{tabular}}}
    \label{tab:detection}
\end{table*}

\begin{table*}[]
    \centering
    \caption{Object detection and instance segmentation results on COCO \textit{val} 2017 with Mask R-CNNs as detectors. AP$^b$ and AP$^m$ are box average precision and mask average precision, respectively.}
    \scalebox{0.99}{
        \begin{tabular}{c|l|rr|lll|lll|r}
            \toprule
            Method & Backbone & \#Params & FLOPs & AP$^b$ & AP$^b_{50}$ & AP$^b_{75}$ & AP$^m$ & AP$^m_{50}$ & AP$^m_{75}$ & FPS  \\
            \midrule
            \multirow{25}{*}{Mask R-CNN} & PVT-M & 63.9M & $-$ & 42.0 & 64.4 & 45.6 & 39.0 & 61.6 & 42.0 & 12 \\
            \multirow{25}{*}{$1\times$ schedule} & Swin-T & 47.8M & 264.0G & 42.2 & 64.6 & 46.2 & 39.1 & 61.6 & 42.0 & \textbf{20} \\
            & Twins-PCPVT-S & 44.3M & 245.0G & 42.9 & 65.8 & 47.1 & 40.0 & 62.7 & 42.9 & 11 \\
            & TransCNN-B & 46.4M & $-$ & 44.0 & 66.4 & 48.5 & 40.2 & 63.3 & 43.2 & 5 \\
            & ViL-M & 60.1M & 261.1G & 43.3 & 65.9 & 47.0 & 39.7 & 62.8 & 42.0 & 5\\
            & RegionViT-B & 92.2M & 287.9G & 43.5 & 66.7 & 47.4 & 40.1 & 63.4 & 43.0 & 8 \\
            & RegionViT-B+ & 93.2M & 307.1G   & 44.5 & 67.6 & 48.7 & 41.0 & 64.4 & 43.9 & 10 \\
            & PVTv2-B2 & 45.0M & $-$ & 45.3 & 67.1 & 49.6 & 41.2 & 64.2 & 44.4 & 14 \\
            & ELSA-T & 49.0M & 269.0G & 45.6 & 67.9 & 50.3 & 41.1 & 64.8 & 44.0 & $-$ \\
            & ScalableViT-S & 46.0M & 256.0G & 45.8 & 67.6 & 50.0 & 41.7 & 64.7 & 44.8 & 12 \\
            & \textbf{CrossFormer-S} & 50.2M & 301.0G & 45.4 & 68.0 & 49.7 & 41.4 & 64.8 & 44.6 & 13 \\
            & \textbf{CrossFormer-S$^\ddagger$} & 50.2M & 291.1G & 45.0 & 67.9 & 49.1 & 41.2 & 64.6 & 44.3 & 14 \\
            & \textbf{CrossFormer++-S$^\ddagger$} & 43.0M & 287.4G & \textbf{46.4}\,\textcolor{blue}{(+1.4)} & 68.8 & 51.3	& \textbf{42.1}\,\textcolor{blue}{(+0.9)} & 65.7 & 45.4 & 17 \\
            \cmidrule{2-11}
            & CAT-B & 71.0M & 356.0G & 41.8 & 65.4 & 45.2 & 38.7 & 62.3 & 41.4 & 13 \\
            & PVT-L & 81.0M & 364.0G & 42.9 & 65.0 & 46.6 & 39.5 & 61.9 & 42.5 & 9 \\
            & Twins-SVT-B & 76.3M & 340.0G & 45.1 & 67.0 & 49.4 & 41.1 & 64.1 & 44.4 & 14 \\
            & ViL-B & 76.1M & 365.1G & 45.1 & 67.2 & 49.3 & 41.0 & 64.3 & 44.2 & 4 \\
            & Twins-SVT-L & 119.7M & 474.0G & 45.2 & 67.5 & 49.4 & 41.2 & 64.5 & 44.5 & 7 \\
            & Swin-S & 69.1M & 354.0G & 44.8 & 66.6 & 48.9 & 40.9 & 63.4 & 44.2 & \textbf{14} \\
            & Swin-B & 107.2M & 496.0G & 45.5 & $-$ & $-$ & 41.3 & $-$ & $-$ & 10 \\
            & PVTv2-B4 & 82.2M & $-$ & 47.5 & 68.7 & 52.0 & 42.7 & 66.1 & 46.1 & 7 \\
            & ScalableViT-B & 95.0M & 349.0G & 46.8 & 68.7 & 51.5 & 42.5 & 65.8 & 45.9 & 9\\
            & \textbf{CrossFormer-B} & 71.5M & 407.9G & 47.2 & 69.9 & 51.8 & 42.7 & 66.6 & 46.2 & 9\\
            & \textbf{CrossFormer-B$^\ddagger$} & 71.5M & 398.1G & 47.1 & 69.9 & 52.0 & 42.7 & 66.5 & 46.1 & 9\\
            & \textbf{CrossFormer++-B$^\ddagger$} &  71.5M	& 408.0G & \textbf{47.7}\,\textcolor{blue}{(+0.6)} & 70.2 & 52.7 & \textbf{43.2}\,\textcolor{blue}{(+0.5)} & 67.3 & 46.7 & 12 \\
            \midrule
            
            \multirow{16}{*}{Mask R-CNN} & PVT-M & 63.9M & $-$ & 44.2 & 66.0 & 48.2 & 45.0 & 63.1 & 43.5 & 12 \\
            \multirow{16}{*}{$3\times$ schedule} & ViL-M & 60.1M & 261.1G & 44.6 & 66.3 & 48.5 & 40.7 & 63.8 & 43.7 & 5 \\
            & Swin-T & 47.8M & 264.0G & 46.0 & 68.2 & 50.2 & 41.6 & 65.1 & 44.8 & \textbf{20} \\
            & Shuffle-T & 48.0M & 268.0G & 46.8 & 68.9 & 51.5 & 42.3 & 66.0 & 45.6 & \textbf{20} \\
            & ViTAEv2-S & 37.0M & $-$ & 47.8 & 69.4 & 52.2 & 42.6 & 66.6 & 45.8 & 11 \\
            & ScalableViT-S & 46.0M & 256.0G & 48.7 & 70.1 & 53.6 & 43.6 & 67.2 & 47.2 & 12 \\
            & \textbf{CrossFormer-S$^\ddagger$}& 50.2M & 291.1G & 48.7 & 70.7 & 53.7 & 43.9 & 67.9 & 47.3 & 14\\
            & \textbf{CrossFormer++-S$^\ddagger$}& 43.0M & 287.4G & \textbf{49.5}\,\textcolor{blue}{(+0.8)} & 71.6 & 54.1	& \textbf{44.3}\,\textcolor{blue}{(+0.4)} & 68.5 & 47.6 & 17 \\
            \cmidrule{2-11}
            & PVT-L & 81.0M & 364.0G & 44.5 & 66.0 & 48.3 & 40.7 & 63.4 & 43.7  & 9 \\
            & ViL-B & 76.1M & 365.1G & 45.7 & 67.2 & 49.9 & 41.3 & 64.4 & 44.5 & 4 \\
            & Shuffle-S & 69.0M & 359.0G & 48.4 & 70.1 & 53.5 & 43.3 & 67.3 & 46.7 & 13 \\
            & Swin-S & 69.1M & 354.0G & 48.5 & 70.2 & 53.5 & 43.3 & 67.3 & 46.6 & \textbf{14} \\
            & ScalableViT-B &  95.0M & 349.0G & 49.0 & 70.3 & 53.6 & 43.8 & 67.4 &  47.5 & 9 \\
            & Shunted-B & 59.0M & $-$ & 50.1 & 70.9 & 54.1 & 45.2 & 68.0 & 48.0 & 4 \\
            & \textbf{CrossFormer-B$^\ddagger$} & 71.5M & 398.1G & 49.8 & 71.6 & 54.9 & 44.5 & 68.8 & 47.9 & 9 \\
            & \textbf{CrossFormer++-B$^\ddagger$} & 71.5M & 408.0G	& \textbf{50.2}\,\textcolor{blue}{(+0.4)} & 71.8 & 54.9 & \textbf{44.6}\,\textcolor{blue}{(+0.1)} & 68.7 & 48.1 & 12 \\
            
            \bottomrule
    \end{tabular}}
    \label{tab:detection-2}
\end{table*}

\subsection{Image Classification}

\textbf{Experimental Settings.} The experiments on image classification are conducted on the ImageNet dataset. It contains 1.28M natural images for training and 50,000 images for evaluation. The images are resized to $224\times 224$ for both training and evaluation by default. The same training settings as the other vision transformers are adopted. In particular, we use an AdamW~\cite{DBLP:journals/corr/KingmaB14} optimizer training for 300 epochs with a cosine decay learning rate scheduler, and 20 epochs of linear warm-up are used. The batch size is 1,024 split on 8 V100 GPUs. An initial learning rate of 0.001 and a weight decay of 0.05 are used. Besides, we use drop path rate of $0.1, 0.2, 0.3, 0.5$ for CrossFormer-T, CrossFormer-S, CrossFormer-B, and CrossFormer-L, respectively. As well, the drop path rates of $0.2, 0.3, 0.5, 0.7$ are used for CrossFormer++-S, CrossFormer++-B, CrossFormer++-L, and CrossFormer++-H, respectively. Further, similar to Swin~\cite{DBLP:journals/corr/abs-2103-14030}, RandAugment~\cite{DBLP:conf/nips/CubukZS020}, Mixup~\cite{DBLP:conf/iclr/ZhangCDL18}, Cutmix~\cite{DBLP:conf/iccv/YunHCOYC19}, random erasing~\cite{DBLP:conf/aaai/Zhong0KL020}, and stochastic depth~\cite{DBLP:conf/eccv/HuangSLSW16} are used for data augmentation.

\textbf{Results.} The results are shown in TABLE~\ref{tab:classification}. CrossFormer outperforms all other contemporaneous vision transformers. In specific, compared against strong baselines DeiT, PVT, and Swin, our CrossFormer outperforms them at least absolute $1.2\%$ in accuracy on small models. Compared with CrossFormer, CrossFormer++ brings about 0.8$\%$ accuracy improvement in average. For example, CrossFormer-B achieves 83.4$\%$ in accuracy, while CrossFormer++-B reaches 84.2$\%$ with a negligible extra computational budget. Moreover, CrossFormer++ outperforms all existing vision transformers with similar parameters and a comparable computational budget and throughput.

\subsection{Object Detection and Instance Segmentation}

\textbf{Experimental Settings.} The experiments on object detection and instance segmentation are both done on the COCO 2017 dataset~\cite{DBLP:conf/eccv/LinMBHPRDZ14}, which contains $118$K training and $5$K val images.
We use MMDetection-based~\cite{mmdetection} RetinaNet~\cite{DBLP:journals/pami/LinGGHD20} and Mask R-CNN~\cite{DBLP:conf/iccv/HeGDG17} as the object detection and instance segmentation head, respectively.
For both tasks, the backbones are initialized with the weights pre-trained on ImageNet. Then the whole models are trained with batch size $16$ on $8$ V100 GPUs, and an AdamW optimizer with an initial learning rate of $1 \times 10^{-4}$ is used. Following previous works, we adopt $1\times$ training schedule (\ie, the models are trained for $12$ epochs) when taking RetinaNets as detectors, and images are resized to 800 pixels for the short side. While for Mask R-CNN, both $1\times$ and $3 \times$ training schedules are used. It is noted that multi-scale training~\cite{DBLP:conf/eccv/CarionMSUKZ20} is also employed when taking $3 \times$ training schedules.

\textbf{Results.} The results on RetinaNet and Mask R-CNN are shown in TABLE~\ref{tab:detection} and TABLE~\ref{tab:detection-2}, respectively. As we can see, CrossFormer outperforms most existing vision transformers and shows higher superiority than on image classification task. We think it is because CrossFormer utilizes cross-scale features explicitly, and cross-scale features are particularly vital for dense prediction tasks like objection detection and instance segmentation. However, there are still some other architectures that perform better than CrossFormer, \eg, PVTv2-B3 achieves $0.2\%$ AP higher than CrossFormer-S. Instead, CrossFormer++ surpasses CrossFormer by at least $0.5\%$ AP and outperforms all existing methods. Further, its performance gain over the other architectures gets sharper when enlarging the model, indicating that CrossFormer++ enjoys greater potentials. For example, CrossFormer++-S outperforms ScalableViT-S by 0.8 in AP (48.7 vs. 49.5) when using Mask R-CNN as the detection head, while the CrossFormer++-B outperforms ScalableViT-B by 1.2 in AP.

\begin{table*}[]
    \caption{Semantic segmentation results on the ADE20K validation set. ``MS IOU'' means testing with variable input size.}
    \begin{subtable}[t]{0.45\textwidth}
        \centering
        \setlength{\tabcolsep}{2mm}{
        \scalebox{0.9}{
             \begin{tabular}{l|rrlr}
                \toprule
                \multicolumn{5}{c}{Semantic FPN ($80$K iterations)} \\
                Models & Parameters & FLOPs & IOU & FPS \\
                \midrule
                PoolFormer-S36 & 34.6M & $-$ & 42.0 & 6 \\
                Twins-SVT-B & 60.4M & 261.0G & 45.0 & 11 \\
                Swin-S & 53.2M & 274.0G & 45.2  & 15\\
                ScalableViT-S & 30.0M & 174.0G & 44.9 & 15 \\
                Shunted-S & 26.1M & 183.0G & 48.2 & 7 \\
                \textbf{CrossFormer-S} & 34.3M & 220.7G & 46.0 & 14 \\
                \textbf{CrossFormer-S$^\ddagger$} & 34.3M & 209.8G & 46.4 & 16 \\
                \textbf{CrossFormer++-S$^\ddagger$} & 27.1M & 199.5G & \textbf{47.4}\,\textcolor{blue}{(+1.0)} & \textbf{17} \\
                \midrule
                PoolFormer-M36 & 59.8M & $-$ & 42.4 & 4 \\
                PVTv2-B3 & 49.0M & 249.6G & 47.3 & 10 \\
                ScalableViT-B & 79.0M & 270.0G & 48.4 & 11 \\
                \textbf{CrossFormer-B} & 55.6M & 331.0G & 47.7 & 9 \\
                \textbf{CrossFormer-B$^\ddagger$} & 55.6M & 320.1G & 48.0 & 10 \\
                \textbf{CrossFormer++-B$^\ddagger$} & 55.6M & 331.1G & \textbf{48.6}\,\textcolor{blue}{(+0.6)} & \textbf{12} \\
                \midrule
                Twins-SVT-L & 103.7M & 397.0G & 45.8 & 8 \\
                PVTv2-B5 & 85.7M &  364.4G & 48.7 & 7 \\
                ScalableViT-L& 105.0M & 402.0G & 49.4 & 7 \\
                \textbf{CrossFormer-L} & 95.4M & 497.0G & 48.7 & 6 \\
                \textbf{CrossFormer-L$^\ddagger$} & 95.4M & 482.7G & 49.1 & 7 \\
                \textbf{CrossFormer++-L$^\ddagger$} & 95.5M & 482.8G & \textbf{49.5}\,\textcolor{blue}{(+0.4)} & 8 \\
                \midrule
                 \textbf{CrossFormer++-H$^\ddagger$} & 99.4M & 612.3G & \textbf{49.7} & \textbf{6} \\
                  \bottomrule
        \end{tabular}
        }}
    \end{subtable}
    \hfill
    \begin{subtable}[t]{0.55\textwidth}
        \centering
        \setlength{\tabcolsep}{2mm}{
        \scalebox{0.9}{
            \begin{tabular}{l|rrllr}
                \toprule
                \multicolumn{6}{c}{UPerNet ($160$K iterations)} \\
                Models & Parameters & FLOPs & IOU & MS IOU & FPS \\
                \midrule
                % S3-T & 60.0MM & 954.0G & 44.9 & 46.3 & $-$ \\
                ELSA-Swin-T & 61.0M & 946.0G & $-$ & 47.7 & $-$ \\
                MPViT-S & 52.0M & 943.0G & $-$ & 48.3 & 6 \\
                ScalableViT-S & 57.0M & 931.0G & 48.5 & 49.4 & 7 \\
                Shunted-S & 52.0M & 940.0G & 48.9 & 49.9 & 5 \\
                \textbf{CrossFormer-S} & 62.3M & 979.5G & 47.6 & 48.4 & 7 \\
                \textbf{CrossFormer-S$^\ddagger$} & 62.3M & 968.5G & 47.4 & 48.2 & 7 \\
                \textbf{CrossFormer++-S$^\ddagger$} &53.1M & 963.5G &\textbf{49.4}\,\textcolor{blue}{(+2.0)} & \textbf{50.0}\,\textcolor{blue}{(+1.8)} & \textbf{8} \\
                \midrule
                Swin-S & 81.0M & 1038.0G & 47.6 & 49.5 & \textbf{7} \\
                % S3-S & 81.0MM & 1071.0G & 48.0 & 49.3 & $-$ \\
                ELSA-Swin-S & 85.0M & 1046.0G & $-$ & 50.3 & $-$ \\
                ViTAEv2-S & 49.0M & $-$ & 45.0 & 48.0 & 6 \\
                MPViT-B & 105.0M & 1186.0G &  $-$ & 50.3 & 4 \\
                ScalableViT-B & 107.0M & 1029.0G & 49.5 & 50.4 & 6 \\
                \textbf{CrossFormer-B} & 83.6M & 1089.7G  & 49.7 & 50.6 & 5 \\
                \textbf{CrossFormer-B$^\ddagger$} & 83.6M & 1078.8G  & 49.2 & 50.1 & 6 \\
                \textbf{CrossFormer++-B$^\ddagger$} & 83.7M &1089.8G & \textbf{50.7}\,\textcolor{blue}{(+1.5)}& \textbf{51.0}\,\textcolor{blue}{(+0.9)} & 6 \\
                \midrule
                Swin-B & 121.0M & 1088.0G & 48.1 & 49.7 & \textbf{6} \\
                ScalableViT-L & 135.0M & 1162.0G & 49.8 & 50.7 & 5 \\
                \textbf{CrossFormer-L} & 125.5M & 1257.8G & 50.4 & 51.4 & 4 \\
                \textbf{CrossFormer-L$^\ddagger$} & 125.5M & 1243.5G & 50.5 & 51.4 & 4 \\
                \textbf{CrossFormer++-L$^\ddagger$} & 125.5M & 1257.9G & \textbf{51.0}\,\textcolor{blue}{(+0.5)} & \textbf{51.9}\,\textcolor{blue}{(+0.5)} & 5 \\
                \midrule
                \textbf{CrossFormer++-H$^\ddagger$} & 129.5M & 1416.1G & \textbf{51.2} & \textbf{51.8} & \textbf{4} \\
                \bottomrule
        \end{tabular}
        }}
    \end{subtable}
    \label{tab:segmentation}
\end{table*}

\subsection{Semantic Segmentation}

\textbf{Experimental Settings.} ADE20K~\cite{DBLP:conf/cvpr/ZhouZPFB017} is used as the benchmark for semantic segmentation. It covers a broad range of $150$ semantic categories, including $20$K images for training and $2$K for validation. Similar to models for detection, we initialize the backbones with weights pre-trained on ImageNet, and MMSegmentation-based~\cite{mmseg2020} semantic FPN and UPerNet~\cite{DBLP:conf/eccv/XiaoLZJS18} are taken as the segmentation head. For FPN~\cite{DBLP:conf/cvpr/KirillovGHD19}, we use an AdamW optimizer with learning rate and weight deacy of $1\times10^{-4}$. Models are trained for $80$K iterations with batch size $16$. For UPerNet, an AdamW optimizer with an initial learning rate of $6\times10^{-5}$ and a weight decay of $0.01$ is used, and models are trained for $160$K iterations. %  with batch size $16$.

\textbf{Results.} All results are shown in TABLE~\ref{tab:segmentation}. Compared with architectures released earlier, CrossFormer exhibits a greater performance gain over the others when enlarging the model, which is similar to object detection. For example, CrossFormer-T achieves $1.4\%$ absolutely higher in IOU than Twins-SVT-B, but CrossFormer-B achieves $3.1\%$ absolutely higher in IOU than Twins-SVT-L. Totally, CrossFormer shows a more significant advantage over the others on dense prediction tasks (\eg, detection and segmentation) than on classification, implying that cross-scale interactions in the attention module are more important for dense prediction tasks than for classification.

Moreover, CrossFormer++ shows a more significant advantage than CrossFormer when using a more powerful segmentation head. As we can see in TABLE~\ref{tab:segmentation}, UPerNet is a more powerful model than semantic FPN. CrossFormer++S outperforms CrossFormer-S by 1.0 AP when using semantic FPN (47.4 vs. 46.4), while outperforms 2.0 AP when using UPerNet as the segmentation head.

\subsection{Ablation Studies}

\begin{table}[t]
    \centering
    \caption{Ablation studies on different kernel sizes of CELs. The results are tested on the ImageNet validation set and the COCO instance segmentation task, and the baseline model is CrossFormer-S (82.5\%) with Mask R-CNN being the segmentation head (41.4 AP).}
    \scalebox{1.0}{
        \setlength{\tabcolsep}{0.9mm}{
            \begin{tabular}{l|l|l|l|crr}
                \toprule
                \multicolumn{4}{c|}{CEL's Kernel Size} & \multirow{2}{*}{\#Params/FLOPs} & \multirow{2}{*}{Acc.} &  \multirow{2}{*}{AP$^m$}\\
                \textit{Stage-1} & \textit{Stage-2} & \textit{Stage-3} & \textit{Stage-4} & & & \\
                \midrule
                $4$ & $2$ & $2 $ & $2$ & 28.3M / 4.5G & 81.5\% & 39.7 \\
                $8$ & $2$ & $2$ & $2$ & 28.3M / 4.5G & 81.9\% & 40.2 \\
                $4, 8$ & $2, 4$ & $2, 4$ & $2, 4$ & 30.6M / 4.8G & 82.3\% & $-$ \\
                % $4, 8, 16, 32$ & $2, 4$ & $2, 4$ & $2, 4$ & 30.8M / 5.1G & \textbf{82.4\%} & $-$\\
                $4, 8, 16, 32$ & $2, 4$ & $2, 4$ & $2, 4$ & 30.7M / 4.9G & \textbf{82.5\%} & \textbf{41.4}\\
                $4, 8, 16, 32$ & $2, 4, 8$ & $2, 4$ & $2, 4$ & 30.9M / 5.1G & 82.3\% & $-$ \\
                $4, 8, 16, 32$ & $2, 4, 8$ & $2, 4$ & $2$ & 29.4M / 5.0G & 82.4\% & $-$ \\
                \bottomrule
        \end{tabular}}
    }
    \label{tab:apd-classification}
\end{table}

\subsubsection{Cross-scale Tokens vs. Single-scale Tokens}

We conduct the experiments by replacing cross-scale embedding layers with single-scale ones.
As we can see in TABLE~\ref{tab:apd-classification}, when using single-scale embeddings, the $8 \times 8$ kernel in \textit{Stage-1} brings $0.4\%$ ($81.9\%$ vs. $81.5\%$) absolute improvement compared with the $4 \times 4$ kernel. It tells us that overlapping receptive fields help improve the model's performance. Besides, all models with cross-scale embeddings perform better than those with single-scale embeddings. In particular, our CrossFormer achieves $1\%$ ($82.5\%$ vs. $81.5\%$) absolute performance gain compared with using single-scale embeddings for all stages. For cross-scale embeddings, we also try several different combinations of kernel sizes, and they all show similar performance ($82.3\% \sim 82.5\%$). In summary, cross-scale embeddings can bring a large performance gain, yet the model is relatively robust with respect to different choices of kernel size.

In addition, we also test different dimension allocation schemes for CEL. As described in Fig.~\ref{fig:embedding}, we allocate different number of dimensions for different sampling kernels, and we also compare with ``allocating equally''. The experiments are conducted with CrossFormer-S and the results are in Table~\ref{tab:cel-allocation}. The results indicate that the two allocation schemes achieve similar accuracy while our scheme owns less parameters.

\begin{table}[t]
    \centering
    \caption{Ablation studies on the dimension allocation scheme for CEL. Wherein, CEL-Equally means allocating the dimension equally for all sampling kernels.}
    \scalebox{1.0}{
        \setlength{\tabcolsep}{1mm}{
            \begin{tabular}{l|c|cccc|rr|r}
                \toprule
                \multirow{2}{*}{CEL Type}& \multicolumn{5}{c|}{Stage-1 Kernels} & \multirow{2}{*}{Parameters} & \multirow{2}{*}{FLOPs} & \multirow{2}{*}{Accuracy} \\
                % \cmidrule{2-6}
                % \line
                & Size & 4 & 8 & 16 & 32 & & & \\
                \midrule
                CEL-Equally & \multirow{2}{*}{Dim.} & 24 & 24 & 24 & 24 & 30.8M & 5.1G & 82.4\% \\
                CEL-Ours & & 48 & 24 & 12 & 12 & \textbf{30.7M} & \textbf{4.9G} & \textbf{82.5\%} \\
                \bottomrule
        \end{tabular}
        }
    }
    \label{tab:cel-allocation}
\end{table}

\begin{table}[t]
\caption{Ablation studies on different self-attention mechanisms.
The results are tested on the ImageNet validation set, and the baseline model is CrossFormer-S (82.5\%), \ie, LSDA with CEL and DPB.}
\centering
\setlength{\tabcolsep}{5mm}{
\begin{tabular}{l|cc|c}
    \toprule
    Self-attentions & CEL & DPB  &  Accuracy \\
    \midrule
    CAT & \checkmark & \checkmark & 81.1\%  \\
    PVT & \checkmark &  & 81.3\%  \\
    LSDA & & & 81.5\%  \\
    Swin & \checkmark & \checkmark & 81.9\% \\
    DaViT & \checkmark & \checkmark & 82.1\%  \\
    CSwin & \checkmark & \checkmark & \textbf{82.5\%}  \\
    ScalableViT & \checkmark & \checkmark & \textbf{82.5\%}  \\
    LSDA &  \checkmark & \checkmark & \textbf{82.5\%}  \\
    \bottomrule
\end{tabular}}
\label{tab:ablation-1}
\end{table}

\subsubsection{LSDA vs. Other Self-attentions}

Self-attention mechanisms used in other vision transformers are also compared. Specifically, we replace LSDA in CrossFormer-S with self-attention modules proposed in previous work, and CEL and DPB are retained (except for PVT because DPB cannot be applied to PVT). As shown in TABLE~\ref{tab:ablation-1}, the self-attention mechanisms have great influences on the models' accuracies. In particular, LSDA outperforms CAT-like self-attention by 1.4\% (82.5\% vs. 81.1\%). Besides, some self-attention mechanisms can achieve similar accuracy to LSDA (CSwin and ScalableViT), which indicates that the most proper self-attention mechanism is not unique.

\begin{table}[t]
       \renewcommand\arraystretch{1.0}
    \centering
    \caption{Results on the ImageNet validation set. We test with different group sizes and ACL Layers. $x \rightarrow y$ indicates linearly scaling from $x$ to $y$.}
    \scalebox{1.0}{\setlength{\tabcolsep}{0.9mm}{
            \begin{tabular}{l|c|cccc|c}
                \hline
                \multirow{2}{*}{Models} & \multirow{2}{*}{ACL} & \multicolumn{4}{c|}{Group Size} & \multirow{2}{*}{Acc.} \\ 
                & & Stage-1 & Stage-2 & Stage-3 & Stage-4 & \\
                \hline
                CrossFormer-S$_1$ & & $4 \times 4$ & $4 \times 4$ & $7 \times 7$ & $7 \times 7$ & 82.4\% \\
                CrossFormer-S$_2$ & & $7 \times 7$ & $7 \times 7$ & $7 \times 7$ & $7 \times 7$ & 82.5\% \\
                CrossFormer-S$_3$& & $4 \times 4$ & $4 \times 4$ & $14 \times 14$ & $7 \times 7$ & 82.9\% \\
                \hline
                % CrossFormer++-S$_1$ & \checkmark & $7 \times 7$ & $7 \times 7$ & $7 \times 7$ & $7 \times 7$ & 82.5\% \\
                CrossFormer++-S$_1$ & \checkmark & $4 \times 4$ & $4 \times 4$ & $7 \times 7$ & $7 \times 7$ & 82.6\% \\
                CrossFormer++-S$_2$ & \checkmark & \multicolumn{3}{c}{$4 \times 4 \rightarrow 14 \times 14$} & $7 \times 7$ & 82.9\% \\
                % CrossFormer++-S$_3$ & \checkmark & $4 \times 4$ & $4 \times 4$ & $7 \times 7$ / $14 \times 14$ & $7 \times 7$ & 83.0\% \\
                CrossFormer++-S$_3$ & \checkmark &  $7 \times 7$ & $7 \times 7$ & $14 \times 14$ & $7 \times 7$ & \textbf{83.2\%} \\
                CrossFormer++-S$_4$ & \checkmark & $4 \times 4$ & $7 \times 7$ & $14 \times 14$ & $7 \times 7$ & \textbf{83.2\%} \\
                CrossFormer++-S & \checkmark & $4 \times 4$ & $4 \times 4$ & $14 \times 14$ & $7 \times 7$ & \textbf{83.2\%} \\
                \hline
                CrossFormer-B & & $4 \times 4$ & $4 \times 4$ & $14 \times 14$ & $7 \times 7$ & 83.9\% \\
                CrossFormer++-B & \checkmark & $4 \times 4$ & $4 \times 4$ & $14 \times 14$ & $7 \times 7$ & 84.2\% \\
                \hline
                CrossFormer-L & & $4 \times 4$ & $4 \times 4$ & $14 \times 14$ & $7 \times 7$ & 84.3\% \\
                CrossFormer++-L & \checkmark & $4 \times 4$ & $4 \times 4$ & $14 \times 14$ & $7 \times 7$ & 84.7\% \\
                \hline
                Swin-T & & $7 \times 7$ & $7 \times 7$ & $7 \times 7$ & $7 \times 7$ & 81.3\% \\
                Swin-T$_1$ & & $4 \times 4$ & $4 \times 4$ & $14 \times 14$ & $7 \times 7$ & 82.8\% \\
                Swin-T$_2$ & \checkmark & $4 \times 4$ & $4 \times 4$ & $14 \times 14$ & $7 \times 7$ & 83.2\% \\
                \hline
    \end{tabular}}}
    \label{sec3:tab:ablation}
\end{table}

\subsubsection{Ablation Studies about PGS and ACL}
% also with Swin results
\textbf{Manually Designed Group Size.} We adopt manually designed group size for the CrossFormer++, \ie, [4, 4, 14, 7] for four stages, respectively. We also test other group size, and the results are shown in TABLE~\ref{sec3:tab:ablation}. Compare CrossFormer++-S with CrossFormer++-S$_3$ and CrossFormer++-S$_4$, and the results show that a $7 \times 7$ group size for the first two stages is inessential, which leads to a greater computational budget but no accuracy improvement. The comparison between CrossFormer++-S and CrossFormer++-S$_1$ (83.2\% vs 82.6\%) shows that a large group size for Stage-3 is pivotal. These conclusions also coincide with our observation that the self-attention at shallow layers concentrates on a small region around each token, while the attention gradually disperses at deep layers.

\textbf{Linearly Scaling Group Size.} Additionally, we also try a simple non-manually designed group size, \ie, we expand group size from $4 \times 4$ to $14 \times 14$ linearly from Stage-1 to Stage-3. However, the linear group size does not work as well as a manually designed group size. 

\textbf{Attention Mechanisms vs. Group Size.} It is also worth emphasizing that the conclusions about group size also apply to other vision transformers. As shown in TABLE~\ref{sec3:tab:ablation}, an appropriate group size for Swin-T also brings a significant accuracy improvement, from 81.3\% to 82.8\%. In comparison, replacing shifted window in Swin with LSDA only brings 0.6\% improvement (as shown in TABLE~\ref{tab:ablation-1}), which indicates that an appropriate group size may be more important than the choice of the self-attention mechanism.

% \subsubsection{Ablation Studies about ACL}
\textbf{Ablation Studies about ACL.} Regarding ACL, the ACL layer brings about 0.3\%$\sim$0.4\% accuracy improvement. Concretely, CrossFormer++-B improves from 83.9\% to 84.2\% after plugging an ACL, and CrossFormer++-L improves from 84.3\% to 84.7\%. Moreover, ACL is a universal layer that also applies to other vison transformers. For example, plugging an ACL into Swin-T can bring 0.4\% accuracy improvement (82.8\% vs. 83.2\%).

\subsubsection{DPB vs. Other Position Representations}

\begin{table}[t]
\centering
\caption{Comparisons between different position representations. The base model is CrossFormer-S. Flexibility indicates whether the representation applies to dynamic group size.}
\setlength{\tabcolsep}{1mm}{
\begin{tabular}{lcccc}
    \toprule
    Method& \#Params/FLOPs & Throughput & Flexibility & Acc. \\
    \midrule
    APE & 30.9M/4.9G & 686 imgs/sec& & 82.1\% \\
    RPB & 30.7M/4.9G & 684 imgs/sec& & \textbf{82.5\%} \\
    DPB & 30.7M/4.9G & 672 imgs/sec & \checkmark & \textbf{82.5\%} \\
    DPB-residual & 30.7M/4.9G & 672 imgs/sec &\checkmark & 82.4\% \\
    \bottomrule
\end{tabular}}
\label{tab:ablation-3}
\end{table}

\begin{table}[t]
    \centering
    \caption{Comparisons between DBP and offline/online interpolated RPB. The baseline is CrossFormer-S fine-tuned with Mask R-CNNs as detectors and evaluated at COCO \textit{val} 2017.}
    \scalebox{0.88}{
        \setlength{\tabcolsep}{0.8mm}{
            \begin{tabular}{l|l|lll|lll}
                \toprule
                Methods  & Throughput & AP$^b$ & AP$^b_{50}$ & AP$^b_{75}$ & AP$^m$ & AP$^m_{50}$ & AP$^m_{75}$ \\   
                \midrule
                % CrossFormer-S-RPB & - &  \\
                Offline interpolated RPB & \textbf{16} imgs/sec & 43.9 & 67.0 & 48.1 & 40.7 & 63.7 & 43.9 \\
                Online interpolated RPB & 14 imgs/sec & 44.5 & 67.4 & 48.8 & 41.0 & 64.2 & 44.3 \\
                DPB & 14 imgs/sec & \textbf{45.4} & \textbf{68.0} & \textbf{49.7} & \textbf{41.4} & \textbf{64.8} & \textbf{44.6} \\
                \bottomrule
            \end{tabular}
        }
    }
    \label{tab:DPB-abl}
\end{table}

We compare the parameters, FLOPs, throughputs, and accuracies of the models among absolute position embedding (APE), relative position bias (RPB), and DPB. The results are shown in TABLE~\ref{tab:ablation-3}. DPB-residual means DPB with residual connections. Both DPB and RPB outperform APE with absolute $0.4\%$ accuracy, which indicates that relative position representations are more beneficial than absolute ones.

Further, DPB achieves the same accuracy ($82.5\%$) as RPB with an ignorable extra cost; however, as we described in Sec.~\ref{sec:dpb}, it is more flexible than RPB and applies to variable image size or group size.
Besides, the results also show that residual connection in DPB does not help improve or even degrades the model's performance (82.5\% vs. 82.4\%).

Moreover, we design two interpolation strategies to adapt RPB to variable group size, called offline interpolated RPB and online interpolated RPB:
\begin{itemize}
    \item Offline interpolated RPB: After training the model on the ImageNet with a small group size (\eg, $14 \times 14$), we first interpolate RPB to a sufficient large size (\eg, applying to at most $41 \times 41$ group size for the COCO dataset). During the fine-tuning process, the enlarged RPB is fine-tuned along the whole model. For the inference stage on downstream tasks, the enlarged RPB is fixed and no interpolation is required.
    \item Online interpolated RPB: The RPB is kept as the original (\eg, fit to $14 \times 14$ group size) after training. During the fine-tuning process, we dynamically resize RPB with a differentiable bilinear interpolation for each image. Though kept as a fixed small size, the RPB is fine-tuned to apply to variable group size. For the inference stage on downstream tasks, the RPB needs to do an online interpolation for each image.
\end{itemize}
As object detection and instance segmentation are the most common tasks with variable input image sizes, experiments are done on the COCO dataset, and results are shown in Table~\ref{tab:DPB-abl}. As we can see, DPB outperforms both offline and online interpolated DPB. Offline interpolated RPB is 1.5 lower than DPB in absolute AP$^b$. We assume it is because there are too many different group sizes ($41 \times 41$), resulting in each of them has limited training samples, so the interpolated RPB may be underfitting. In contrast, online interpolation RPB encodes performs a little better than the offline version, but it is still not as profitable as DPB. Moreover, since online interpolation RPB executes an online interpolation for each image, its throughput is on par with DPB, which is a bit smaller than that of offline interpolated RPB (16 imgs/sec vs. 14 imgs/sec). The throughput indicates that the extra computational budgets for online interpolated RPB and DPB are both acceptable.

\section{Conclusions and Future Works} \label{sec:6}
In this paper, we first proposed a universal visual backbone, dubbed CrossFormer. It utilizes cross-scale features through our proposed CEL and LSDA. The experimental results show that utilizing cross-scale features explicitly can significantly improve the vision transformers' performance on image classification and other downstream tasks. Besides, a more flexible relation position representation, DPB, is also proposed to make the CrossFormer apply to dynamic group size. Based on CrossFormer, we further analyzed the self-attention module's and MLPs' outputs in the CrossFormer and proposed progressive group size (PGS) and amplitude cooling layer (ACL). Based on PGS and ACL, CrossFormer++ achieves better performance than CrossFormer and outperforms all the existing state-of-the-art vision transformers on representative visual tasks. Besides, extensive experiments show that CEL, PGS, and ACL are all universal and can consistently bring performance gains when plugged into other vision transformers.

Despite these contributions, the proposed modules still have some limitations. Concretely, though we demonstrated the importance of a progressive appropriate group size, an empirically designed group size is used finally. An adaptive and automated group size policy is still expected. Then, to prevent the gradient vanishing issue, the ACL layer cannot be used in large quantities in vision transformers. So, we will also explore how to cool down the vision transformers' amplitude without cutting the residual connection.

\ifCLASSOPTIONcompsoc
  \section*{Acknowledgments}
  This work was supported in part by The National Nature Science Foundation of China (Grant Nos: 62036009, 62273302, 62273303, 62303406, 61936006, U1909203), in part by Ningbo Key R\&D Program (No.2023Z231, 2023Z229), in part by the Key R\&D Program of Zhejiang Province, China (2023C01135), in part by Yongjiang Talent Introduction Programme (Grant No: 2022A-240-G, 2023A-194-G), in part by the Key Research and Development Program of Zhejiang Province (No. 2021C01012).
\else
  \section*{Acknowledgment}
\fi


\ifCLASSOPTIONcaptionsoff
  \newpage
\fi



\bibliographystyle{IEEEtran}
\bibliography{CrossFormer}


\begin{IEEEbiography}[{\includegraphics[width=1in,height=1.25in,clip,keepaspectratio]{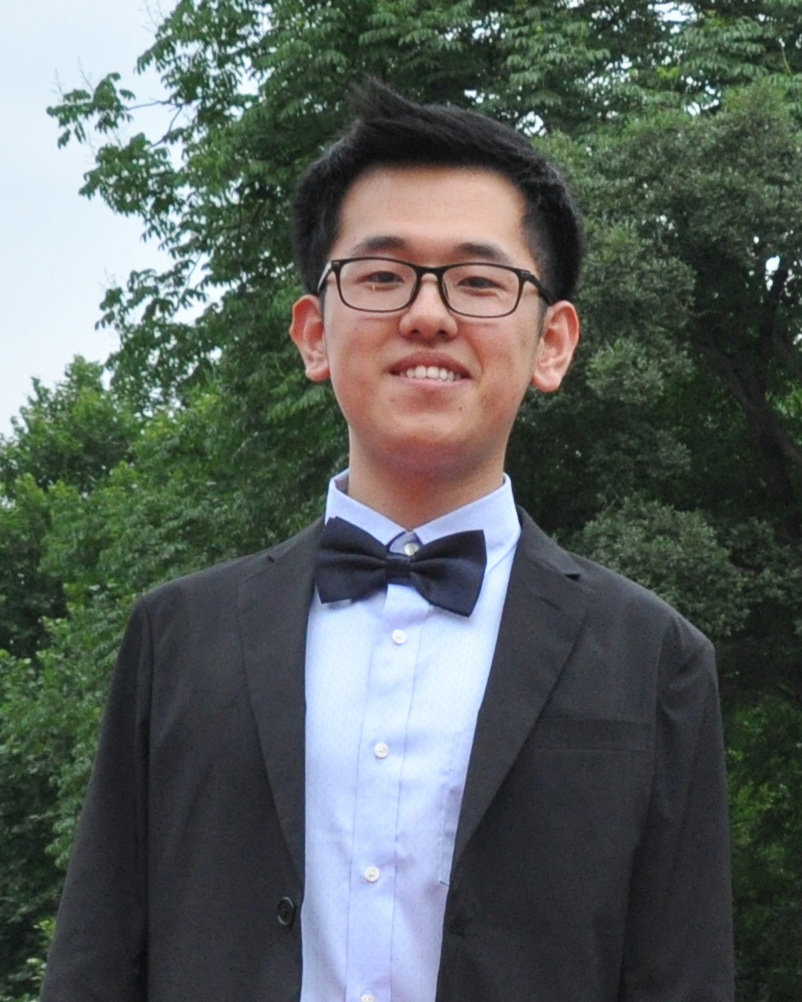}}]{Wenxiao Wang} is a distinguished research fellow, School of Software Technology of Zhejiang University, China. He received the Ph.D. degree in computer science and technology from Zhejiang University in 2022. His research interests include deep learning and computer vision.
\end{IEEEbiography}

\begin{IEEEbiography}[{\includegraphics[width=1in,height=1.25in,clip,keepaspectratio]{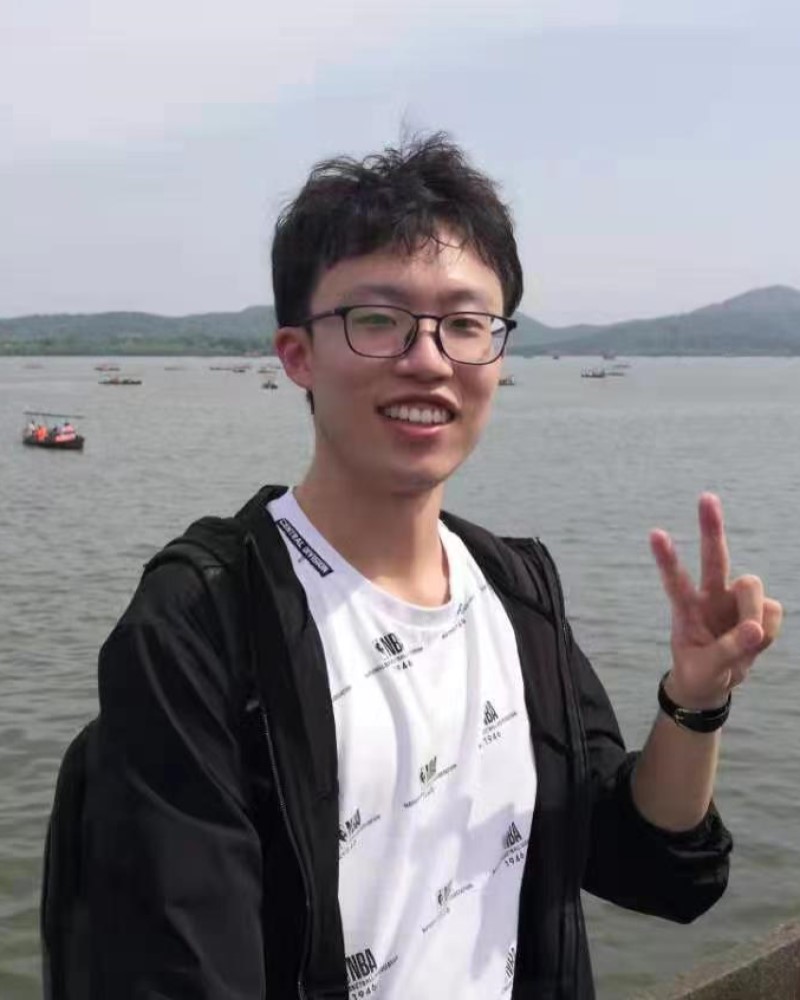}}]{Wei Chen} is a senior student from College of Computer Science and Technology at Zhejiang University, China. He majors in Artificial Intelligence and is currently an intern in the State Key Lab of CAD\&CG at Zhejiang University. His research interests include deep learning and computer vision.
\end{IEEEbiography}

\begin{IEEEbiography}[{\includegraphics[width=1in,height=1.25in,clip,keepaspectratio]{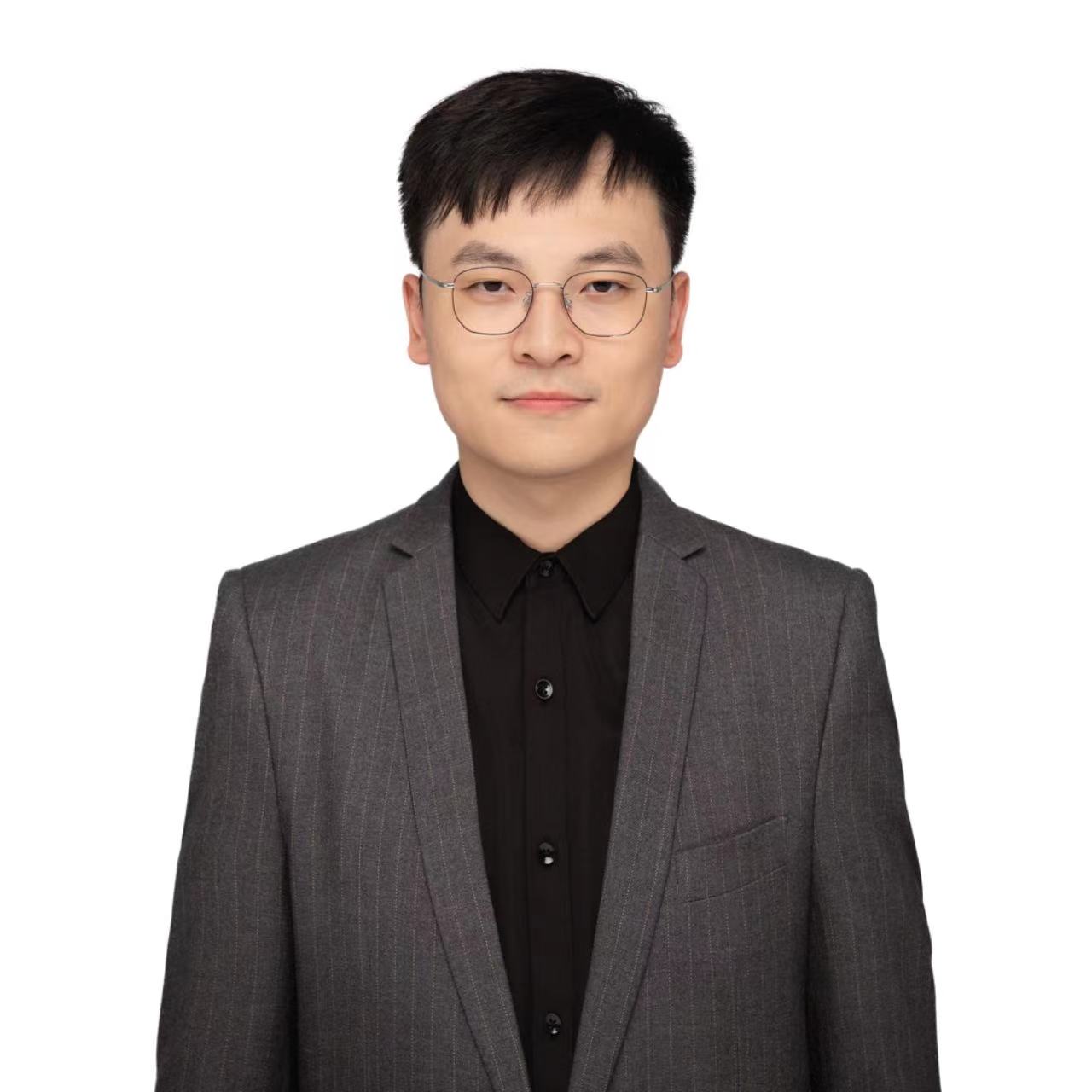}}]{Qibo Qiu} received the M.S degree in computer science at Zhejiang University, Hangzhou, China, in 2017. He is currently a senior algorithm engineer with Zhejiang Laboratory, his research interests include road marker detection, 2D/3D lane detection, general place recognition, localization and perception for intelligent vehicles and mobile robots.
\end{IEEEbiography}

\begin{IEEEbiography}
[{\includegraphics[width=1in,height=1.25in,clip,keepaspectratio]{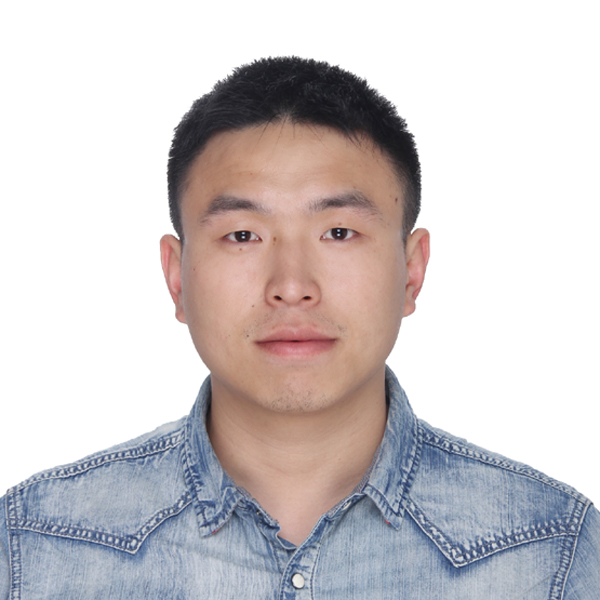}}]{Long Chen} received the Ph.D. degree in Computer Science from Zhejiang University in 2020, and the B.Eng. degree in Electrical Information Engineering from Dalian University of Technology in 2015. He is currently an assistant professor at the Department of Computer Science and Engineering, HKUST. He was a postdoctoral research scientist at Columbia University and a senior researcher at Tencent AI Lab. 
His research interests are computer vision and multimedia.
\end{IEEEbiography}

\begin{IEEEbiography}[{\includegraphics[width=1in,height=1.25in,clip,keepaspectratio]{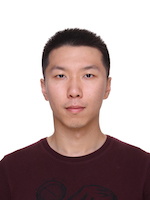}}]{Boxi Wu} is a distinguished research fellow at the School of Software Technology, Zhejiang University. He received his bachelor and Ph.D. degree in computer science and technology from Zhejiang University in 2016 and 2022. His research interests include machine learning and computer vision.
\end{IEEEbiography}

\begin{IEEEbiography}[{\includegraphics[width=1in,height=1.25in,clip,keepaspectratio]{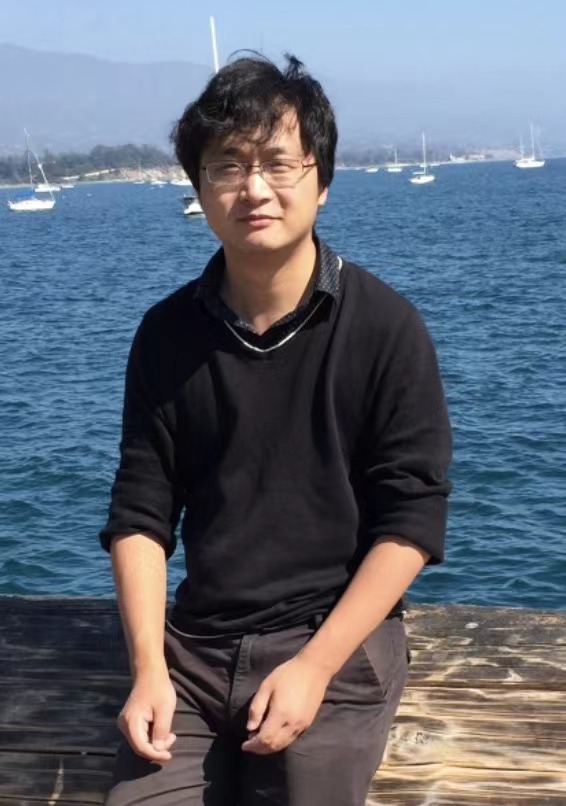}}]{Binbin Lin} is an assistant professor in the School of Software Technology at Zhejiang University, China. He received a Ph.D degree in computer science from Zhejiang University in 2012. His research interests include machine learning and decision making.
\end{IEEEbiography}

\begin{IEEEbiography}[{\includegraphics[width=1in,height=1.25in,clip,keepaspectratio]{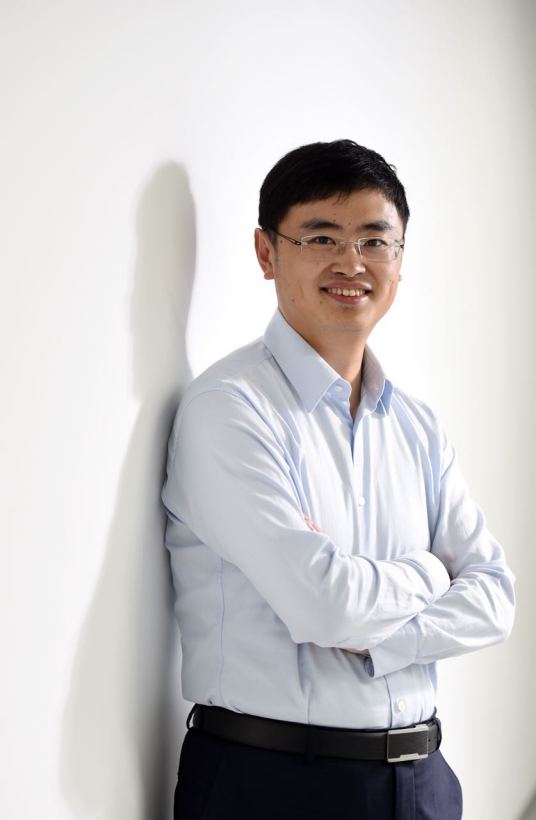}}] {Xiaofei He} is currently a professor in the State Key Lab of CAD\&CG at Zhejiang University, and the CEO of FABU Technology Co., Ltd. His research interest includes machine learning, deep learning, and autonomous driving. He has authored/co-authored more than 200 technical papers with Google Scholar Citation over 38,000 times. He received the best paper award from AAAI, 2012. He is a fellow of IAPR and a senior member of IEEE.
\end{IEEEbiography}

\begin{IEEEbiography}[{\includegraphics[width=1in,height=1.25in,clip,keepaspectratio]{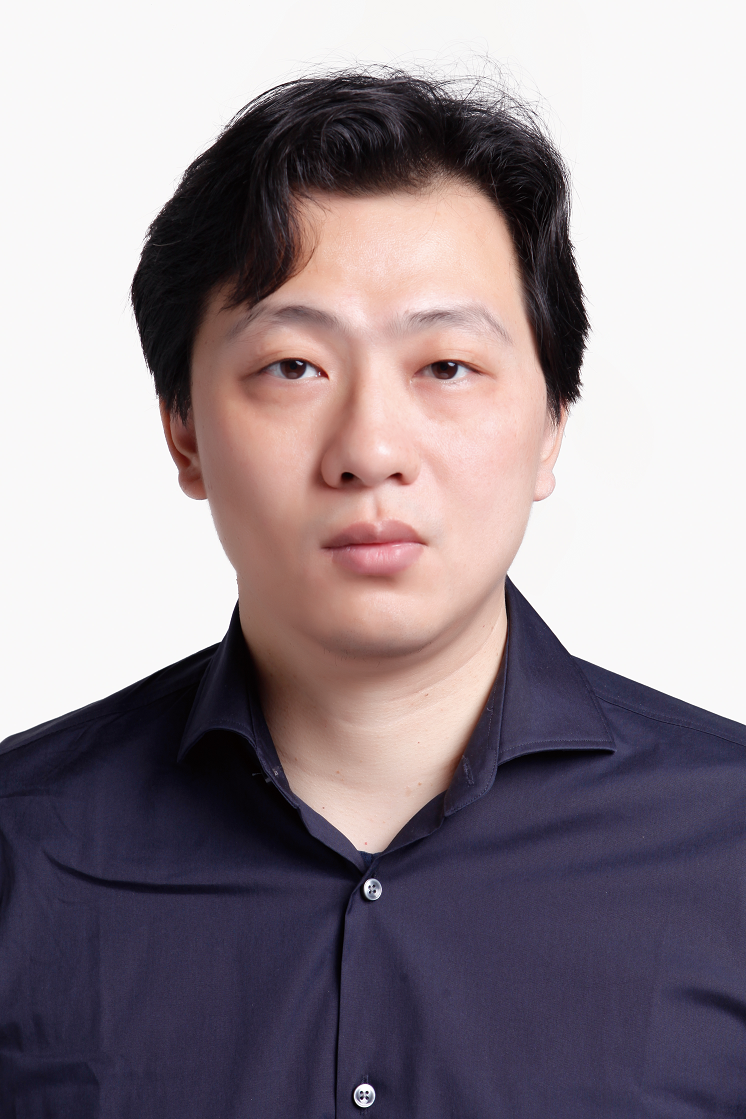}}]{Wei Liu} (M'14-SM'19-F'23) received the Ph.D. degree from Columbia University in 2012 in Electrical Engineering and Computer Science. He is currently a Distinguished Scientist of Tencent and the Director of Ads Multimedia AI at Tencent Data Platform. Prior to that, he has been a research staff member of IBM T. J. Watson Research Center, USA from 2012 to 2015. Dr. Liu has long been devoted to fundamental research and technology development in core fields of AI, including deep learning, machine learning, computer vision, pattern recognition, information retrieval, big data, etc. To date, he has published extensively in these fields with more than 280 peer-reviewed technical papers, and also issued 32 US patents. He currently serves on the editorial boards of IEEE TPAMI, TNNLS, and IEEE Intelligent Systems. He is an Area Chair of top-tier computer science and AI conferences, e.g., NeurIPS, ICML, IEEE CVPR, IEEE ICCV, IJCAI, and AAAI. Dr. Liu is a Fellow of the IEEE, IAPR, and IMA.   
\end{IEEEbiography}


%\newpage



%\vfill


\end{document}

\title{Appendices for ``CrossFormer++: A Versatile Vision Transformer Hinging on Cross-scale Attention''}

% \author{Wenxiao Wang,
%         Wei Chen,
%         Qibo Qiu,
%         Long Chen,
%         Boxi Wu,
%         Binbin Lin, \\
%         Xiaofei He,~\IEEEmembership{Senior Member,~IEEE,}
%         and Wei Liu,~\IEEEmembership{Fellow,~IEEE}
% % \IEEEcompsocitemizethanks{
% % \IEEEcompsocthanksitem Wei Liu is with Data Platform, Tencent.}%
% % \thanks{Manuscript received January 31, 2023.}
% }

\newcommand\ie{\textit{i.e.}}
\newcommand\eg{\textit{e.g.}}

\maketitle

\appendices

\begin{table*}[]
    \centering
    \caption{CrossFormer-based backbones for object detection and semantic/instance segmentation. The example input size is $1280 \times 800$. $D$ and $H$ mean embedding dimension and the number of heads in the multi-head self-attention module, respectively. $G$ and $I$ are group size and interval for SDA and LDA, respectively.}
    \scalebox{1.0}{
        \setlength{\tabcolsep}{3mm}{
            \begin{tabular}{cc|c|cccc}
                \toprule
                & Output Size & Layer Name & CrossFormer-T & CrossFormer-S & CrossFormer-B & CrossFormer-L \\ \midrule
                \multirow{5}{*}{Stage-1} & \multirow{5}{*}{$ 320 \times 200 $} & Cross Embed. & \multicolumn{4}{c}{Kernel size: $4 \times 4$, $8 \times 8$, $16 \times 16$, $32 \times 32$, Stride=$4$} \\
                \cmidrule{3-7} 
                & & \multirow{3}{*}{SDA/LDA} & \multirow{2}{*}{$\begin{bmatrix}D_1=64 \\H_1=2 \\G_1=14 \\I_1=16\end{bmatrix} \times 1$} & \multirow{2}{*}{$\begin{bmatrix}D_1=96 \\H_1=3 \\G_1=14 \\I_1=16\end{bmatrix} \times 2$} & \multirow{2}{*}{$\begin{bmatrix}D_1=96\\H_1=3\\G_1=14\\I_1=16, \end{bmatrix} \times 2$} & \multirow{2}{*}{$\begin{bmatrix}D_1=128 \\ H_1=4 \\ G_1=14 \\ I_1=16\end{bmatrix} \times 2$} \\
                &  & \multirow{3}{*}{MLP} &  &  &  \\
                &  &  & & & & \\
                &  &  & & & & \\
                \midrule
                \multirow{5}{*}{Stage-2} & \multirow{5}{*}{$ 160\times100 $} & Cross Embed. & \multicolumn{4}{c}{Kernel size: $2 \times 2$, $4 \times 4$, Stride=$2$} \\ 
                \cmidrule{3-7} 
                & & \multirow{3}{*}{SDA/LDA} & \multirow{3}{*}{$\begin{bmatrix}D_2=128\\H_2=4 \\ G_2=14\\I_2= 8 \end{bmatrix} \times 1$} & \multirow{2}{*}{$\begin{bmatrix}D_2=192\\H_2=6 \\ G_2=14\\I_2= 8 \end{bmatrix} \times 2$} & \multirow{2}{*}{$\begin{bmatrix}D_2=192\\H_2=6 \\ G_2=14\\I_2=8 \end{bmatrix} \times 2$} & \multirow{2}{*}{$\begin{bmatrix}D_2=256\\H_2=8 \\ G_2=14\\I_2=8 \end{bmatrix} \times 2$} \\
                &  & \multirow{3}{*}{MLP} &  &  &  \\
                &  &  & & & & \\
                &  &  & & & & \\
                \midrule
                \multirow{5}{*}{Stage-3} & \multirow{5}{*}{$ 80\times50 $} & Cross Embed. & \multicolumn{4}{c}{Kernel size: $2 \times 2$, $4 \times 4$, Stride=$2$} \\ 
                \cmidrule{3-7} 
                & & \multirow{3}{*}{SDA/LDA} & \multirow{3}{*}{$\begin{bmatrix}D_3=256\\H_3=8 \\ G_3=7\\I_3=2 \end{bmatrix} \times 8$} & \multirow{2}{*}{$\begin{bmatrix}D_3=384\\H_3=12 \\ G_3=7\\I_3=2 \end{bmatrix} \times 6$} & \multirow{2}{*}{$\begin{bmatrix}D_3=384\\H_3=12 \\ G_3=7\\I_3=2 \end{bmatrix} \times 18$} & \multirow{2}{*}{$\begin{bmatrix}D_3=512\\H_3=16 \\ G_3=7\\I_3=2 \end{bmatrix} \times 18$} \\
                &  & \multirow{3}{*}{MLP} &  &  &  \\ 
                &  &  & & & & \\
                &  &  & & & & \\
                \midrule
                \multirow{5}{*}{Stage-4} & \multirow{5}{*}{$ 40 \times 25 $} & Cross Embed. & \multicolumn{4}{c}{Kernel size: $2 \times 2$, $4 \times 4$, Stride=$2$} \\ 
                \cmidrule{3-7}
                & & \multirow{3}{*}{SDA/LDA} & \multirow{3}{*}{$\begin{bmatrix}D_4=512\\H_4=16 \\ G_4=7\\I_4=1 \end{bmatrix} \times 6$} & \multirow{2}{*}{$\begin{bmatrix}D_4=768\\H_4=24 \\ G_4=7\\I_4=1 \end{bmatrix} \times 2$} & \multirow{2}{*}{$\begin{bmatrix}D_4=768\\H_4=24 \\ G_4=7\\I_4=1 \end{bmatrix} \times 2$} & \multirow{2}{*}{$\begin{bmatrix}D_4=1024\\H_4=32 \\ G_4=7\\I_4=1 \end{bmatrix} \times 2$} \\
                &  & \multirow{3}{*}{MLP} &  &  &  \\ 
                &  &  & & & & \\
                &  &  & & & & \\
                \bottomrule
    \end{tabular}}}
    \label{tab:variants-2}
\end{table*}

\section{Efficient Dynamic Position Bias (DPB)} \label{apd:dpb}

\begin{figure}[]
    \centering
    \includegraphics[width=0.6\linewidth]{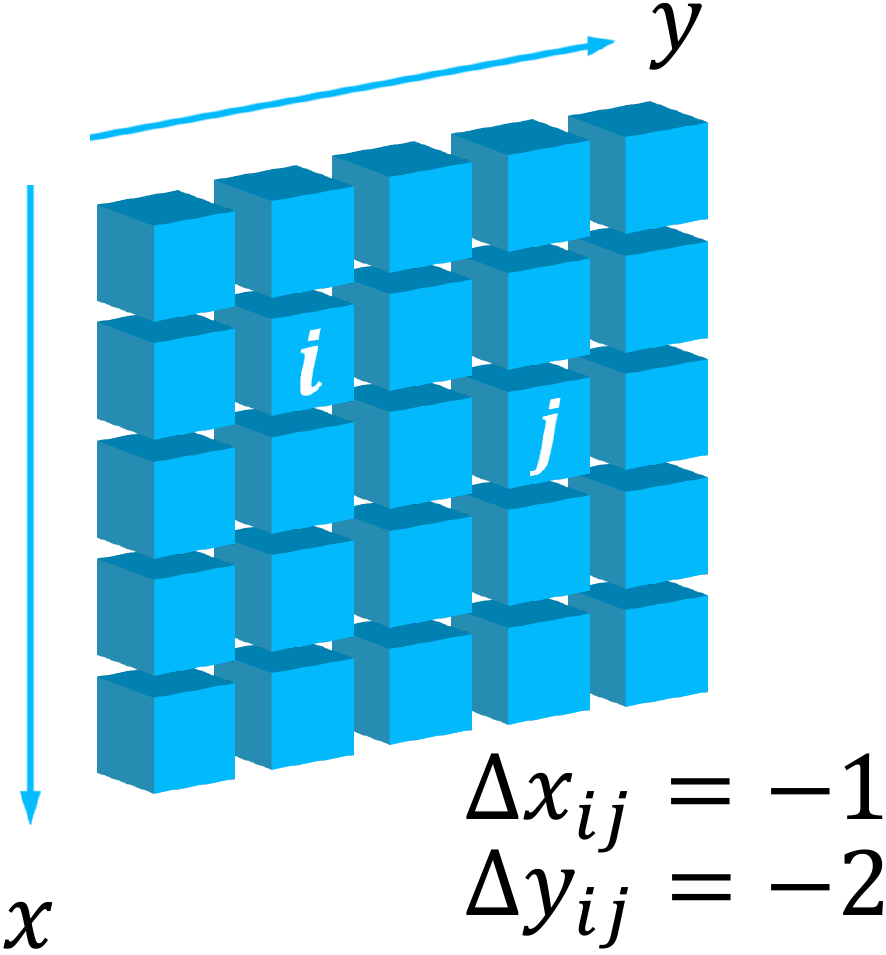}
    \caption{An example of computing $(\Delta x_{ij}, \Delta y_{ij})$ for DPB.}
    \label{fig:effi_DPB}
\end{figure}

Figure~\ref{fig:effi_DPB} gives an example of computing $(\Delta x_{ij}, \Delta y_{ij})$ with $G=5$ in the DPB module. For a group of size $G \times G$, it is easy to deduce that:
\begin{equation}
\label{equ:effi_DPB}
\begin{aligned}
0 &\le x, y < G \\
1 - G &\le \Delta x_{ij} \le G - 1 \\
1 - G &\le \Delta y_{ij} \le G - 1.
\end{aligned}
\end{equation}
Thus, motivated by the relative position bias, we construct a matrix $\hat{B} \in \mathbb{R}^{(2G-1) \times (2G-1)}$, where
\begin{equation}
\label{equ:effi_DPB-2}
\begin{aligned}
\hat{B}_{i, j} = DPB(1-G+i, 1-G+j),\ 0 \le i, j < 2G-1.
\end{aligned}
\end{equation}
The complexity of computing $\hat{B}$ is $O(G^2)$. Then, the bias matrix $B$ in DPB can be drawn from $\hat{B}$, \ie,
\begin{equation}
\label{equ:effi_DPB-3}
\begin{aligned}
B_{i, j} = \hat{B}_{\Delta x_{ij}, \Delta y_{ij}}.
\end{aligned}
\end{equation}
When the image/group size (\ie, $G$) is fixed, both $\hat{B}$ and $B$ will be also unchanged in the test phase. Therefore, we only need to compute $\hat{B}$ and $B$ once, and DPB is equivalent to relative position bias in this case.

\section{Variants of CrossFormer for Detection and Segmentation} \label{apd:variants}
We test two different backbones for dense prediction tasks. The variants of CrossFormer for  dense prediction (object detection, instance segmentation, and semantic segmentation) are in Table~\ref{tab:variants-2}. The architectures are the same as those for image classification except that different $G$ and $I$ in the first two stages are used. Notably, group size (\ie, $G$ and $I$) does not affect the shape of weight tensors, so backbones pre-trained on ImageNet can be fine-tuned directly on other tasks even if they use different $G$ and $I$.

We test two different backbones for dense prediction tasks. The variants of CrossFormer for  dense prediction (object detection, instance segmentation, and semantic segmentation) are in Table~\ref{tab:variants-2}. The architectures are the same as those for image classification except that different $G$ and $I$ in the first two stages are used. Notably, group size (\ie, $G$ and $I$) does not affect the shape of weight tensors, so backbones pre-trained on ImageNet can be fine-tuned directly on other tasks even if they use different $G$ and $I$.

\section{Experiments}

\begin{table*}[t]
    \centering
    \caption{Object detection results on COCO \textit{val} 2017. ``Memory'' means the allocated memory per GPU reported by $torch.cuda.max\_memory\_allocated()$. $^\ddagger$ indicates that models use different $(G, I)$ from classification models.}
    \scalebox{1.0}{
        \begin{tabular}{c|l|ccccr|rr|lll}
            \toprule
            Method & Backbone & $G_1$ & $I_1$ & $G_2$ & $I_2$ & Memory & \#Params & FLOPs & AP$^b$ & AP$^b_{50}$ & AP$^b_{75}$   \\
            \midrule
            \multirow{4}{*}{RetinaNet} & CrossFormer-S & 7 & 8 & 7 & 4 & 14.7G & 40.8M & 282.0G & 44.4 & 65.8 & 47.4\\
            \multirow{4}{*}{$1\times$ schedule} & CrossFormer-S$^\ddagger$& 14 & 16 & 14 & 8 & 11.9G & 40.8M & 272.1G & 44.2 & 65.7 & 47.2\\
            \cmidrule{2-12}
            & CrossFormer-B & 7 & 8 & 7 & 4 & 22.8G & 62.1M & 389.0G & 46.2 & 67.8 & 49.5 \\
            & CrossFormer-B$^\ddagger$& 14 & 16 & 14 & 8 & 20.2G & 62.1M & 379.0G & 46.1 & 67.7 & 49.0 \\
            \midrule
            \multirow{4}{*}{Mask-RCNN} & CrossFormer-S & 7 & 8 & 7 & 4 & 15.5G & 50.2M & 301.0G & 45.4 & 68.0 & 49.7 \\
            \multirow{4}{*}{$1\times$ schedule} & CrossFormer-S$^\ddagger$& 14 & 16 & 14 & 8 & 12.7G & 50.2M & 291.1G & 45.0 & 67.9 & 49.1 \\
            \cmidrule{2-12}
            & CrossFormer-B & 7 & 8 & 7 & 4 &23.8G & 71.5M & 407.9G & 47.2 & 69.9 & 51.8  \\
            & CrossFormer-B$^\ddagger$& 14 & 16 & 14 & 8 & 21.0G & 71.5M & 398.1G & 47.1 & 69.9 & 52.0 \\
            \bottomrule
    \end{tabular}}
    \label{tab:apd-detection}
\end{table*}

\begin{table*}[t!]
    \caption{Semantic segmentation results on ADE20K validation set with semantic FPN or UPerNet as heads.}
    \centering
    \scalebox{1.0}{
        \begin{tabular}{l|cccc|rrrr|rrrrr}
            \toprule
            \multirow{2}{*}{Backbone} & \multirow{2}{*}{$G_1$} & \multirow{2}{*}{$I_1$} & \multirow{2}{*}{$G_2$} & \multirow{2}{*}{$I_2$} & \multicolumn{4}{c|}{Semantic FPN ($80$K iterations)} & \multicolumn{5}{c}{UPerNet ($160$K iterations)} \\
            & & & & & Memory & \#Params & FLOPs & IOU & Memory & \#Params & FLOP & IOU & MS IOU \\
            \midrule
            CrossFormer-S & 7 & 8 & 7 & 4 & 20.9G & 34.3M & 220.7G & 46.0 & $-$ & 62.3M & 979.5G & 47.6 & 48.4 \\
            CrossFormer-S$^\ddagger$& 14 & 16 & 14 & 8 & 20.9G & 34.3M & 209.8G & 46.4 & 14.6G & 62.3M & 968.5G & 47.4 & 48.2 \\
            \midrule
            CrossFormer-B & 7 & 8 & 7 & 4 & 14.6G & 55.6M & 331.0G & 47.7 & 15.8G & 83.6M & 1089.7G & 49.7 & 50.6 \\
            CrossFormer-B$^\ddagger$& 14 & 16 & 14 & 8 & 14.6G & 55.6M & 320.1G & 48.0 & 15.8G & 83.6M & 1078.8G & 49.2 & 50.1 \\
            \midrule
            CrossFormer-L & 7 & 8 & 7 & 4 & 25.3G & 95.4M & 497.0G & 48.7 & 18.1G & 125.5M & 1257.8G & 50.4 & 51.4 \\
            CrossFormer-L$^\ddagger$& 14 & 16 & 14 & 8 & 25.3G & 95.4M &  482.7G & 49.1 & 18.1G & 125.5M & 1243.5G & 50.5 & 51.4 \\
            \bottomrule
    \end{tabular}}
    %   \end{subtable}
    \label{tab:apd-segmentation-1}
\end{table*}

\subsection{Object Detection}

Table~\ref{tab:apd-detection} provides more results on object detection with RetinaNet and Mask-RCNN as detection heads. As we can see, a smaller $(G, I)$ achieves a higher AP than a larger one, but the performance gain is marginal. Considering that a larger $(G, I)$ can save more memory cost, we think $(G_1=14, I_1=16, G_2=14, I_2=8)$, which accords with configurations in Table~\ref{tab:variants-2}, achieves a better trade-off between the performance and cost.

\subsection{Semantic Segmentation}

Similar to object detection, we test two different configurations of $(G, I)$ for semantic segmentation's backbones. The results are shown in Table~\ref{tab:apd-segmentation-1}. As we can see, the memory costs of the two configurations are almost the same, which is different from experiments on object detection. Further, when taking semantic FPN as the detection head, CrossFormers$^\ddagger$ show advantages over CrossFormers on both IOU (\eg, $46.4$ vs. $46.0$) and FLOPs (\eg, $209.8$G vs. $220.7$G). When taking UPerNet as the segmentation head, a smaller $(G, I)$ achieves higher performance like object detection.